\pdfoutput=1

\documentclass[11pt]{article}
\usepackage{listings}

\usepackage{acl} 
\usepackage{times}
\usepackage{latexsym}
\usepackage[T1]{fontenc}
\usepackage[utf8]{inputenc}
\usepackage{microtype}
\usepackage{inconsolata}
\usepackage{graphicx}
\usepackage{colortbl}
\usepackage{amsmath}
\usepackage{booktabs}
\usepackage{amssymb}
\usepackage{adjustbox}
\usepackage{pifont}
\newcommand{\cmark}{\ding{51}}
\newcommand{\xmark}{\ding{55}}
\usepackage{bm}
\usepackage{longtable}
\usepackage{multirow}
\usepackage{multicol}
\usepackage[framemethod=TikZ]{mdframed}
\usepackage{xcolor}
\usepackage{url}
\usepackage{enumitem}
\usepackage{verbatim}
\usepackage{algorithm}
\usepackage{algpseudocode}
\usepackage{afterpage}
\usepackage{authblk}
\usepackage{hyperref}

\newcommand{\cofirst}{$^\ast$}
\newcommand{\corresponding}{\textsuperscript{$\dagger$}}
\newcommand{\addnote}{$^\ddagger$}

\title{Mobile-Bench: An Evaluation Benchmark for LLM-based Mobile Agents}

\author[13\cofirst\ \addnote]{\bf Shihan Deng}
\author[13\cofirst\ \addnote]{\bf Weikai Xu}
\author[23\cofirst\ \addnote]{\bf Hongda Sun}
\author[3]{\bf Wei Liu}

\author[2]{\bf Tao Tan}
\author[3]{\bf Jianfeng Liu}
\author[3]{\\\bf Ang Li}
\author[3]{\bf Jian Luan}
\author[3]{\bf Bin Wang}

\author[2\corresponding]{\bf Rui Yan}
\author[1\corresponding]{\bf Shuo Shang}

\affil[1]{University of Electronic Science and Technology of China}
\affil[2]{Gaoling School of Artificial Intelligence, Renmin University of China}
\affil[3]{XiaoMi AI Lab}

\affil[ ]{\texttt{\{dengshihan7, xuwk266, jedi.shang\}@gmail.com}}
\affil[ ]{\texttt{\{sunhongda98, ruiyan\}@ruc.edu.cn}}

\begin{document}

\maketitle
\let\thefootnote\relax\footnotetext{\cofirst\ Equal contribution.}
\let\thefootnote\relax\footnotetext{\corresponding\ Corresponding authors: Rui Yan and Shuo Shang.}
\let\thefootnote\relax\footnotetext{\addnote\ Work done during the internship at XiaoMi.}
\begin{abstract}
With the remarkable advancements of large language models (LLMs), LLM-based agents have become a research hotspot in human-computer interaction.
However, there is a scarcity of benchmarks available for LLM-based mobile agents.
Benchmarking these agents generally faces three main challenges:
(1) The inefficiency of UI-only operations imposes limitations to task evaluation.
(2) Specific instructions within a singular application lack adequacy for assessing the multi-dimensional reasoning and decision-making capacities of LLM mobile agents.
(3) Current evaluation metrics are insufficient to accurately assess the process of sequential actions. 
To this end, we propose Mobile-Bench, a novel benchmark for evaluating the capabilities of LLM-based mobile agents.
First, we expand conventional UI operations by incorporating 103 collected APIs to accelerate the efficiency of task completion.
Subsequently, we collect evaluation data by combining real user queries with augmentation from LLMs.
To better evaluate different levels of planning capabilities for mobile agents, our data is categorized into three distinct groups: SAST, SAMT, and MAMT, reflecting varying levels of task complexity. Mobile-Bench comprises 832 data entries, with more than 200 tasks specifically designed to evaluate multi-APP collaboration scenarios.
Furthermore, we introduce a more accurate evaluation metric, named CheckPoint, to assess whether LLM-based mobile agents reach essential points during their planning and reasoning steps. Dataset and platform are available at \href{https://github.com/XiaoMi/MobileBench}{https://github.com/XiaoMi/MobileBench}.
\end{abstract}

\section{Introduction}
Interacting with mobile devices using natural language is a long-standing pursuit in human-computer interaction ~\citep{bolt1980put,karat2002conversational,folstad2017chatbots}. With the remarkable advancements in large language models (LLM) ~\citep{bai2022constitutional,chowdhery2022palm,du2021glm,touvron2023llama,ouyang2022training}, LLM-driven agents are at the forefront, yet their reasoning capability to navigate mobile application functionalities lags behind their proficiency with web pages on PCs ~\citep{yao2022webshop,sun2023determlr}.  
To faithfully replicate a typical mobile environment, it's imperative to incorporate a diverse set of applications and leverage authentic data, moving beyond the limitations of purely simulated scenarios. The development challenges in the mobile domain stem from a trio of core issues: a limited understanding of mobile interfaces, a scarcity of application variety, and a lack of real-world data. 

\begin{figure}[t]
  \centering
  \includegraphics[width=\linewidth, page=1]{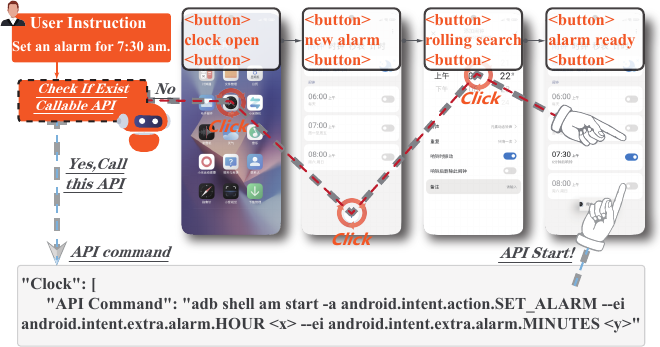}
  \caption{For the task of ``\textit{Setting an alarm for seven thirty.}'', accomplishing it solely through UI operations requires four steps, while API calls can achieve the same task in just one step.}
  \label{fig:model_intro}
\end{figure}

\begin{table*}[h!]
\centering
\renewcommand{\arraystretch}{1.2}
\begin{tabular}{@{}lccccc@{}}
\hline
\textbf{Platform\&BenchMark} & \textbf{InfoUI} & \textbf{API\&UI} & \textbf{Real APP} & \textbf{Real Query} & \textbf{Multi-APP} \\
\hline
World of Bits~\cite{shi2017world} & \cmark & \xmark & \xmark & \xmark & \xmark\\
WebShop~\cite{yao2022webshop} & \cmark & \xmark & \xmark & \xmark & \xmark\\
AndroidEnv~\cite{toyama2021androidenv} & \xmark & \xmark & \cmark & \xmark & \xmark\\
MobileEnv~\cite{zhang2023mobile} & \cmark & \xmark & \cmark & \xmark & \xmark\\
\hline
\textbf{Mobile-Bench (Ours)} & \cmark & \cmark & \cmark & \cmark & \cmark\\
\hline
\end{tabular}
\caption{Comparison of Mobile-Bench with existing LLM-based agent platforms. `InfoUI' represents whether UI information is used for interaction with agents, `API\&UI' represents whether the agent's actions like API calls and UI interface operations, 'Real APP' represents whether real applications are used, `Real Query' represents whether real user queries are used, and `Multi-APP' represents whether there are tasks involving multiple applications.}
\label{tab:benchmark}
\end{table*}

Due to Google's breakthrough \citep{wang2023enabling} in UI interface representation, LLM agent's understanding of UI pages becomes easier, leading to the creation of UI platforms such as Android-Env \cite{toyama2021androidenv} and Mobile-Env \cite{zhang2023mobile}, which tasks are defined within individual games or search engines.
However, these works collectively face the following challenges: 
(1) UI actions depend on the textual descriptions of interfaces, where structured text fails to capture the content of graphical buttons or images which can lead to wrong actions. A single API action might be equivalent to dozens of UI steps, leading to UI's inefficiency.
(2) Their tasks are far removed from real-world task scenarios encountered in daily use, which require cooperation between multiple applications, with user commands being ambiguous and not specifying target applications. 
(3) The evaluation of tasks should not solely rely on LLMs, without any objective quantitative metrics.

In fact, voice assistants on mobile phones can meet most of the users' daily needs, yet they do not interact directly with UI interfaces but operate by invoking the APIs~\cite{qin2023toolllm} behind applications.
As shown in Figure \ref{fig:model_intro}, in mobile applications, APIs are more efficient than UI interfaces; a single API call can be equivalent to multiple UI operations to achieve the same outcome.
However, a single API is insufficient for more complex tasks, especially when user commands are unclear, necessitating reliance on LLMs to interpret user intent. Therefore, an agent capable of utilizing both UI and APIs would be best suited for the job. Simultaneously, It requires developing a strategy for the selection and order of the application usage, with human oversight merely focusing on reviewing the outcomes. This is a function that voice assistants currently lack~\cite{wen2023empowering,wen2023droidbot}.
To this end, we develop a combination of API and UI actions to circumvent the limitations of UI interfaces, each action can be chosen between UI interactions and API calls; all tasks begin from the mobile HOME page rather than from the launch page of a specific application, enabling the agent to determine single or multiple applications it will use; queries in the task are gathered from real users, and instruction generation is only applied to some complex ones which undergo rigorous manual review; we draw inspiration from objective metrics in software automation testing, named CheckPoint, and have made necessary adjustments to accommodate the unpredictable semantic outputs of LLMs.
Above all, we propose a mobile phone environment that includes a platform supporting both API and UI interactions, and a corresponding dataset with multi-APP tasks.
Table \ref{tab:benchmark} presents a comparison among recent platforms and benchmark work based on API and UI.

Our contributions are summarized as follows:\\
 \indent(1) To the best of our knowledge, we are the first to establish a running platform for LLM-based mobile agents that simultaneously supports both UI and API calls.\\
\indent(2) We propose an evaluation dataset containing diverse tasks for multi-APP interactions. Our tasks starting from the home page are more appropriate for testing the planning capabilities for agents. Our dataset and platform will be released soon.\\
\indent(3) We introduce a new category-based evaluation metric to assess the task completion capabilities of the agent in the context of both UI and API interactions. 

\section{Related Work}
\subsection{Mobile Platforms}
Prior to the emphasis on LLM-based agents, research efforts were directed towards RL-based agents, exemplified by the Android-Env platform ~\cite{toyama2021androidenv}. This open-source platform tailored for reinforcement learning experiments within the Android ecosystem, successfully tested various RL-based agents like DDPG ~\cite{zhang2023deep}, D4PG ~\cite{barth2018distributed}, MPO ~\cite{abdolmaleki2018maximum}, DQN ~\cite{mnih2015human}, IMPALA ~\cite{espeholt2018impala} and R2D2 ~\cite{kapturowski2018recurrent}. 

More significant research has focused on LLM-based agents~\cite{liu2024skepticism,sun2024harnessing,sun2024facilitating}. Regarding the domain of tool-using agents, they can be categorized into three main types:

1) For mobile tasks. Platforms like AutoDroid, DroidBot-GPT, GPT-Droid, and WebShop~\cite{wen2023empowering, wen2023droidbot, liu2023chatting, yao2022webshop} create an interactive environment enabling  LLMs to engage with mobile tasks, and generate human-like operations for automation test. Mobile-Env ~\cite{zhang2023mobile} is specifically designed to evaluate agents' capabilities in handling multi-step interactions.

2) For PC Tasks. Researchers developed Toollama ~\cite{qin2023toolllm} to evaluate the capabilities to use tools and API calls. AgentBench ~\cite{liu2023agentbench} presents a standardized Agent task evaluation architecture with strong decoupling and scalability. PPTC Benchmark ~\cite{guo2023pptc} proposed to evaluate the ability of LLM-based agents on PowerPoint tasks.

3) Other Methods. Toolformer ~\cite{schick2023toolformer} and HuggingGPT ~\cite{shen2023hugginggpt} evaluate LLM's capability to master tools. 

\subsection{Benchmarks for LLM agents}\label{sec:cohypo-entanglement}
To assess agents' proficiency in understanding user interfaces, a diverse dataset covering various tasks is crucial~\cite{liu2023agentbench}. The widely used RICO dataset ~\cite{deka2017rico} is commonly employed for this purpose, with Screen2Vec \cite{li2021screen2vec} utilizing it to evaluate agent performance.

However, due to the absence of specific standards for evaluating agent performance, efforts have focused on designing evaluation frameworks. PPTC Benchmark ~\cite{guo2023pptc} devised 279 multi-round dialogue tasks for PPT file operations. DroidTask ~\cite{wen2023empowering} and various unnamed datasets ~\cite{liu2023chatting, wen2023droidbot} covering various mobile applications have also been established. Additionally, Screen2Words used a sampling method to sample screens from the RICO-SCA ~\cite{li2020mapping} dataset and hired professional annotators to generate English summaries for these screens ~\cite{wang2021screen2words}.

Current evaluation standards align with various works. ToolBench proposes Win Rate gauges the model's solution quality against benchmarks like RoBERTa~\cite{liu2019roberta}, GPT-3 ~\cite{brown2020language}, PaLM ~\cite{chowdhery2023palm}, OPT~\cite{zhang2022opt}, ChatGPT ~\cite{bubeck2023sparks} and GPT-4 ~\cite{openai2023gpt4}. Although Fan~\cite{fan2024not} found that the cost of inference can be reduced by using only the necessary layers for inference, it is still expensive to calculate the win rate. Mobile-Env ~\cite{zhang2023mobile} evaluates agent performance based on the completion status, average steps, and average rewards in WikiHow tasks. PPTC Benchmark ~\cite{guo2023pptc} uses Turn-based and Session-based accuracy. Android in the Wild ~\cite{rawles2023android} makes use of Out-of-distribution Generalization. Overall, metrics such as success rate, episode length, and match score are currently the most commonly employed.

\section{Our Environment}
\subsection{Mobile-Bench Benchmark}
\paragraph{Data collection.}
The queries in the dataset are divided into the following three categories: 
\begin{itemize}
    \item \textbf{\textit{SAST: Single-App-Single-Task.}} A real dataset containing only one task text, including single-task operations such as opening and closing APP, such as \textit{"Help me open the map"}.
    \item \textbf{\textit{SAMT: Single-App-Multi-Task.}} A real dataset containing multiple task texts, as well as constructed single-APP data. A complex multi-task on single APP, such as \textit{"Help me open the map, and navigate to Eiffel Tower."}.
    \item \textbf{\textit{MAMT: Multi-App-Multi-Task.}} Constructed multi-APP data, complete a complex multi-task, such as \textit{"Help me search for the latest technology news and share it with friends."}
\end{itemize}
 SAST is directly derived from real voice requests processed by the voice assistants loaded on the mobile phone. We select a subset of this query collection, primarily filtering out the portion that requires voice assistant processing and involves multimodal tools. Additionally, querys that exceed permissions or involve privacy are also filtered out.

\begin{figure}[t]
  \centering
  \includegraphics[width=0.45\textwidth, page=1]{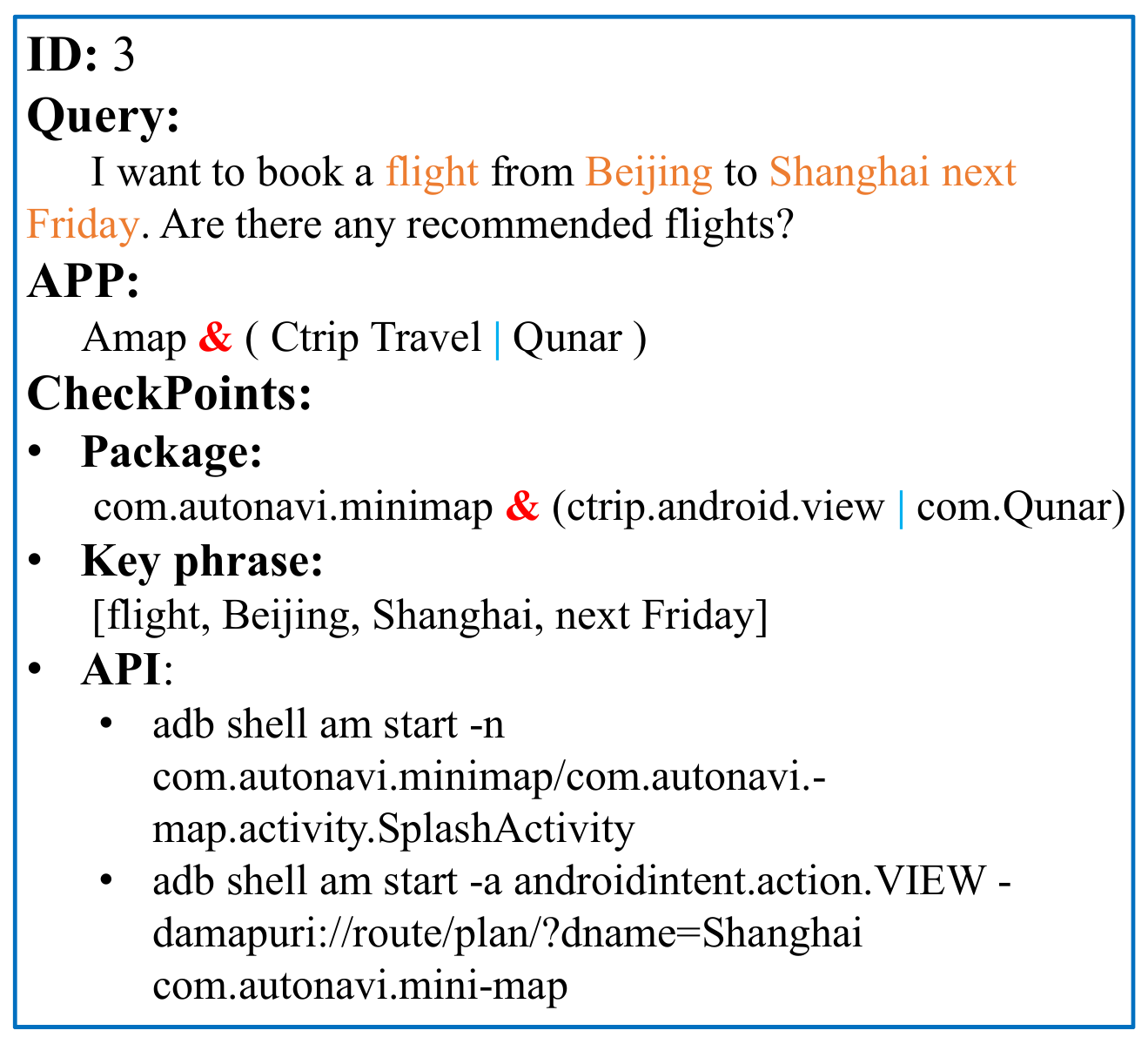}
  \caption{A test case in MAMT. $\&$ stands for conjunction check, CC; $|$ stands for disjunction check, DC; $[ \ ]$ stands for sequential check, SC. The package CheckPoint passes when the action history includes either Amap and Ctrip Travel, or Amap and Qunar. Key phrase CheckPoint comes from the orange parts in the case.}
  \label{fig:model_datacase}
\end{figure}

Since there are fewer SAMT and MAMT data in real data and the quality is not high, refer to Toollama ~\cite{qin2023toolllm} method, we use GPT-4 to construct SAMT and MAMT data. For MAMT, we randomly sample 6 applications from the entire application collection, and then provide some examples of real multi-APP data to prompt GPT-4 to select 2-4 applications to generate tasks. By integrating real and constructed data, we create the final dataset. An example of data is shown in Figure \ref{fig:model_datacase}.


\paragraph{APP \& API collection.}
 To ensure task comprehensiveness, we select not only the applications included in SAST and SAMT but also the most popular free applications from each category in the APP Store. Obtaining the API is to analyze the package of each application to obtain its external reserved interface \cite{desnos2011android}. The advantage of this is that the obtained API is naturally classified for the application. Since the description of the API in the decompilation result is not as detailed as the development document, we use the ADB(Android Debug Bridge) command to verify the feasibility of the API one by one. Owing to its debugging properties, system-level APIs can also be invoked normally, allowing access to functions such as checking the battery status and performing memory cleaning. 
 For more specific application names and categories, please refer to Appendix \ref{APP&API statistics}

 \paragraph{Dataset statistics.}
Including several default applications within the system, we collected a total of 29 applications. For applications, we collected a total of 103 usable APIs, which primarily serve the following functions: system calls, opening pages, closing pages, searching for information, viewing details, and controlling device switches. These functions are summarized into the following main aspects: page switch, details view, broadcast, search. In Table \ref{tab:APIfunction}, we have tabulated the number of APIs and the functional categories covered by APIs, categorized by the type of APP. We organized the available APIs and APP descriptions for each APP, and generated an APP list as the basis for selecting applications, shown in Appendix \ref{APP&API statistics}.

\begin{figure}[t]
    \centering
    \includegraphics[width=1\linewidth]{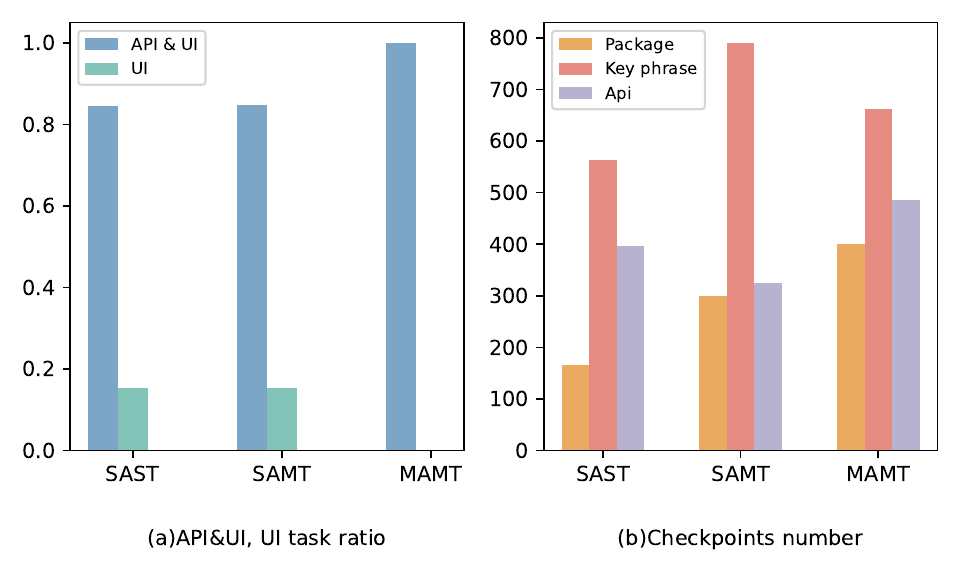}
    \caption{(a) The API$\&$UI, UI task ratio. In SAST and SAMT, API$\&$UI task ratio is $85\%$, in MAMT, it is $100\%$. (b) The number of CheckPoints.}
    \label{fig:image2_bar}
\end{figure}

In the Mobile-Bench dataset, we collected a total of \textit{332, 300, 200} queries for SAST, SAMT, and MAMT. We sort out the APIs actually used by each task in real voice requests. Provide these API as an example to GPT-4 for query generation. As shown in Figure \ref{fig:image2_bar}(a), we calculated the ratio of tasks calling APIs, ensuring a sufficient number of tasks in the dataset that include steps to call API. This approach ensures that we have sufficient data to analyze the role of APIs in task completion.

\begin{table}[t]
\centering
\renewcommand{\arraystretch}{1.2}
\begin{adjustbox}{max width=0.5\textwidth}
\small
\begin{tabular}{@{}lcccc@{}}
\hline
\textbf{APP Category} & \textbf{API Quantity} & \textbf{APP Number} & \textbf{API Functions}\\
 \hline
Travel Transportation & 5 & 3 & \ding{172}, \ding{173}, \ding{175} \\
Audiovisual Vision & 15 & 5 & \ding{172}, \ding{173}, \ding{174} \\
Social Communication & 3 & 1 & \ding{172}, \ding{173}, \ding{175}\\
Fashion Shopping & 14 & 6 & \ding{172}, \ding{175} \\
Information News & 11 & 4 & \ding{172}, \ding{173}, \ding{175} \\
Practical Tool & 38 & 8 & \ding{172}, \ding{173}, \ding{174}, \ding{175}, \ding{176} \\
Home Life & 5 & 1 & \ding{172}, \ding{176} \\
Book Reading & 7 & 2 & \ding{172}, \ding{173}, \ding{175} \\
Universal Buttons & 5 & 0 & \ding{176}\\
\hline
\end{tabular}
\end{adjustbox}
\caption{Our dataset covers nine major categories of applications, and we compared them based on the API function. The above API functions can be summarized into five categories: \ding{172} Page Navigation, \ding{173} Viewing Details, \ding{174} Playback, \ding{175} Searching, and \ding{176} System Calls. }
\label{tab:APIfunction}
\end{table}

\paragraph{Quality verification.}\cite{bolotova2023wikihowqa}
 The initial test data originates from software automation tests, but some complex data points are generated by GPT-4. To ensure the quality of our dataset, we randomly sampled 100 data points from each of the SAST, SAMT, and MAMT, resulting in a total of 300 quality test data. We conducted cross-source validation to verify the feasibility of these CheckPoints. The specific formula for calculation is as follows: 
\begin{equation}
\text{Overlap}(CP_1, CP_2) = \frac{|CP_1 \cap CP_2|}{|CP_1|}
\end{equation}
\textit{CP\textsubscript{1},CP\textsubscript{2}} representing the CheckPoint sequences generated by \textit{CP\textsubscript{instruction}} and \textit{CP\textsubscript{Human}}, respectively. In Table \ref{tab:overlap}, we list the human evaluation results for three types of data. From the table, it can be observed that a higher proportion of terminal data corresponds to better data quality. However, all MAMT data is generated by instructions, its quality does not exhibit an unacceptable gap compared to SAST. See appendix \ref{Dataset quality analysis} for more analysis.
\begin{table}[h!]
\centering
\renewcommand{\arraystretch}{1.2}
\begin{adjustbox}{max width=\textwidth}
\small
\begin{tabular}{@{}lcccc@{}}
\hline
\textbf{Statistics} & \textbf{SAST} & \textbf{SAMT} & \textbf{MAMT} & \textbf{Total}\\
 \hline
 CP\textsubscript{instruction} & 395 & 546 & 513 & 1454 \\
 CP\textsubscript{Human} & 412 & 598 & 623 & 1633 \\
 CP\textsubscript{instruction \! $\cap$ \! Human} & 372 & 466 & 412 & 1250 \\
 \hline
 Overlap & 0.94 & 0.85 & 0.80 & 0.86 \\
\hline
\end{tabular}
\end{adjustbox}
\caption{Human Evaluation Results}
\label{tab:overlap}
\end{table}

\subsection{Test Platform}

\begin{figure}[t]
  \centering
\includegraphics[width=0.40\textwidth]{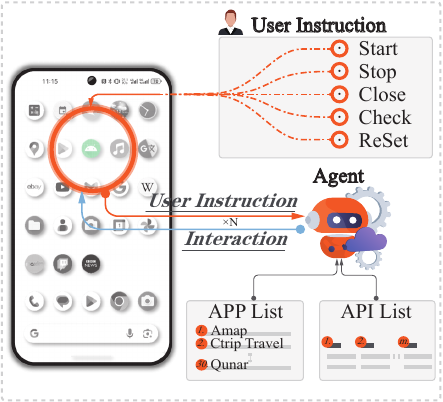}
  \caption{Test Platform Overview. The test platform is linked by the user, the simulator, and the Agent. After the user's instructions are issued, the entire test execution process is completed by the Agent, which can view and manage the test tasks through the preset interface in the cloud.}
  \label{main_model}
\end{figure}

\textbf{Overview}
Mobile-Bench is designed as a universal interaction platform that supports hybrid API and UI interactions. Users are able to construct their own evaluation data following a fixed format, yet they must adhere to our prescribed evaluation method. As shown in Figure \ref{main_model} users can interact with the environment using the following commands. 
\begin{itemize}
\setlength{\itemsep}{0pt}
\setlength{\parsep}{0pt}
\setlength{\parskip}{0pt}
    \item \textbf{\textit{Start}}: Open the \textit{test environment} and load the preset snapshot using this command. Each test case must start from the same environment.
    \item \textbf{\textit{Stop}}: Stop the \textit{test environment} and end test.
    \item \textbf{\textit{Close}}: Close the \textit{test environment} and save the test process and results.
    \item \textbf{\textit{Check}}: Capture a screenshot snapshot of the current \textit{test environment}. 
    \item \textbf{\textit{ReSet}}: Load a previously saved environment snapshot into the \textit{test environment}. 
\end{itemize}

\textbf{Observation space}
To enable the agent to read information on the android emulator in a human-like manner, we use Appium to obtain page information. Following the method described by Wang \citep{wang2023enabling}, we convert XML to HTML, as the training data for LLMs is predominantly sourced from the Internet, which includes numerous HTML files. Therefore, we believe that LLM has a better understanding of HTML than XML. Given the tree structure of XML, we initially convert the XML into a tree format and subsequently transform the nodes that need to be displayed to the agent into HTML. 
The agent simulates human interaction with smartphones, performing three major operations: click, input, and scroll. Humans visually identify which elements can be clicked or receive input, and use their fingers to determine if they can scroll the screen. Therefore, we provide the agent with elements that are visible and scrollable. Due to the limit on context length, we only convert the information required by the agent in XML to HTML:
\begin{itemize}
\setlength{\itemsep}{0pt}
\setlength{\parsep}{0pt}
\setlength{\parskip}{0pt}
    \item \textbf{\textit{Type}}: HTML element categories inherited directly from XML formatted information.
    \item \textbf{\textit{ID}}: ``\textit{ID}'' inherits from the XML ``\textit{resource-id}'' attribute, uniquely identifying the existence of an element. 
	\item \textbf{\textit{Package}}: the package name of the current application.
    \item \textbf{\textit{Class}}: the class of the element, such as \textit{ImageView, TextView}.
    \item \textbf{\textit{Description \& text}}: describe the function and shape of the element.
    \item \textbf{\textit{Clickable \& Scrollable}}: whether the element is clickable and scrollable.
    \item \textbf{\textit{Bounds}}: if the element is scrollable, this attribute will be present and scope the scroll component, such as:
\[\bm{\left[ x_i, y_i \right]\left[ x_j, y_j \right]}\]
The scrollable rectangle ranges from $[x_i,y_i]$ to $[x_j,y_j]$.
\end{itemize}
And, there is an example of HTML elements:
\textit{<button package="com.ximalaya.ting.android" class="android.widget.Button" clickable="true"> message </button>}

\textbf{Action space}
Our Mobile-Bench imitates human behavior in using mobile and summarizes three actions \cite{zhang2023mobile} and imitates the process of calling the API on the test platform \cite{sengupta2023platform}:
\begin{itemize}
\setlength{\itemsep}{0pt}
\setlength{\parsep}{0pt}
\setlength{\parskip}{0pt}
    \item \textbf{\textit{Click}}: simulate real user click actions by passing in specific elements.
    \item \textbf{\textit{Scroll}}: simulate real user scrolling actions by \textit{tapping - dragging - releasing}.
    \item \textbf{\textit{Input}}: simulate real user input actions by \textit{clicking-typing}.
    \item \textbf{\textit{API Call}}: launch an activity or send an intent by invoking an API through ADB commands. 
\end{itemize}
\begin{figure*}[t]
  \centering
\includegraphics[width=\textwidth, page=1]{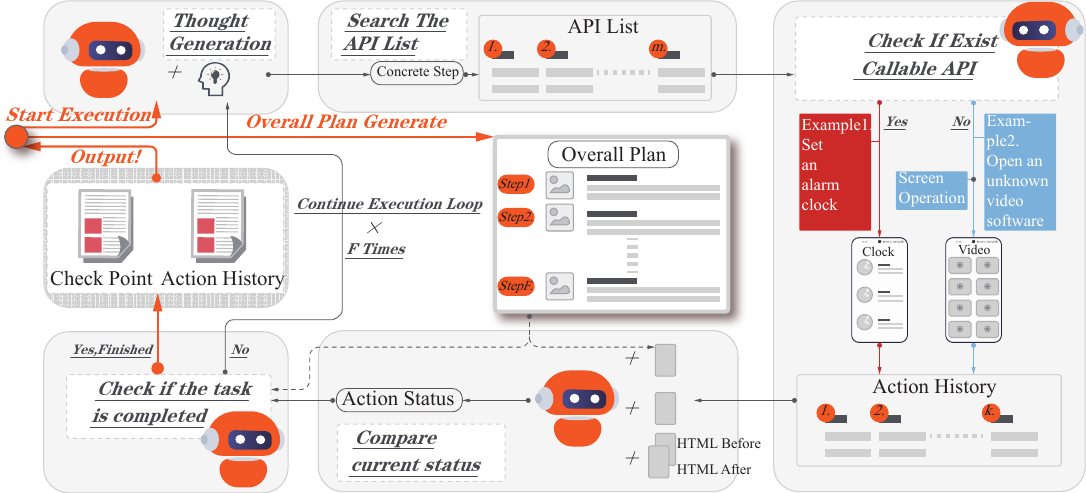}
  \caption{Baseline Model Overview. The entire process framework consists of sensors, reflection components, controllers, execution components, and environments. Once a task starts, these components will run iteratively until the task is completed or the maximum number of steps is reached.}
  \label{in_model}
\end{figure*}
\subsection{Evaluation Method}
\paragraph{CheckPoint.}
Automated test CheckPoint coverage \cite{bajunaid2018efficient} is a test metric for the software execution process. It cannot assist in checking the software results, but it can visually inspect whether the software runs in the specified unit sequence. During data construction, we supply APPs and APIs, which naturally serve as detection indicators. Additionally, we incorporated a CheckPoint to verify if the UI operation correctly clicks on the intended element. After sorting out the above CheckPoints, we constructed the following three CheckPoints: 
\begin{itemize}
\setlength{\itemsep}{0pt}
\setlength{\parsep}{0pt}
\setlength{\parskip}{0pt}
    \item \textbf{\textit{Package}}: the unique package name corresponding to the application. Checking the package can determine whether the correct application is used.
    \item \textbf{\textit{Key phrase}}: the key phrase extracted from the query, represents key steps in the UI execution process.
    \item \textbf{\textit{API}}: API commands that need to be called during the execution process.
\end{itemize}
To evaluate the agent's selection and execution capabilities, we divide the inspection granularity into two levels: \textit{CheckPoint\textsubscript{l1}} - whether it uses the correct application, and \textit{CheckPoint\textsubscript{l2}} - whether it follows the predefined paths to complete the task. For \textit{CheckPoint\textsubscript{l1}}, we check the number of correctly called packages. For \textit{CheckPoint\textsubscript{l2}}, we check the number of correctly called package, key phrase, API.
For CheckPoints, we identify three logical relationships: sequential, conjunctive, and disjunctive checks. These correspond to the instability of LLM output and its tendency for synonym substitution. The calculation formula for "sequential check" is as follows:
\begin{equation}
Score_{Sequen} = \frac{|\sum_{Str \in SC\cap AH} Str |}{|\sum_{Str \in SC} Str |}
\end{equation}
\textit{SC} represent \textit{Sequential Check Set} and \textit{AH} represent \textit{Actions History}. The calculation formulas for conjunctive checks is as follows: 
\begin{equation}
Score_{conjun} = 
\begin{cases} 
1, & if \  \forall str \in CC, str \in AH \\
0, & otherwise
\end{cases}
\end{equation}
\text{CC} represent \textit{Conjunctive Check Set}. The calculation formulas for disjunctive checks is as follows: 
\begin{equation}
Score_{disjun} = 
\begin{cases} 
1, & if \ \exists str  \in  DC  ,  str  \in AH \\
0, & otherwise
\end{cases}
\end{equation}

\textit{DC} represent \textit{Disjunctive Check Set}. The weighted sum of the above three scores will be the final CheckPoint coverage rate. 

As shown in Figure \ref{fig:image2_bar}, the number of key phrase CheckPoints is significantly higher than that of packages, indicating the need for more semantic information to ensure tasks are completed step-by-step. Analyzing the dataset from a proportional perspective, we find that the distributions of the three types of CheckPoints are \textit{{0.212, 0.493, 0.294}}, with key phrase CheckPoints remaining the most predominant method of checking.



In general, a \textit{test case} should include at least the following contents: 
\textbf{\textit{ID, Query, APP List, CheckPoints(Package, Key phrase, API)}}. Figure \ref{fig:model_datacase} is a test case that contains the above three CheckPoints. 


\paragraph{PassRate.}\cite{qin2023toolllm} We assess an agent's human-computer interaction capabilities by calculating the proportion of queries successfully completed within the specified step limits. During this process, we organized the emulator's current state. Subsequently, GPT-4 evaluates the task completion status. We computed the percentage of pass tasks, yielding a PassRate as an indicator of agent's human-computer interaction capabilities.

\paragraph{Average steps.}\cite{zhang2023mobile}
We quantified the step size required by Mobile-Bench to complete tasks as a metric for evaluating the efficiency of the agent. In Mobile-Bench, a 'step' is defined as the completion of a UI operation or the execution of an API call. 

\section{Experiment}
\subsection{Baseline Model}
Our model's architecture, illustrated in Algorithm \ref{alg:example}, begins by obtaining the smartphone's UI information in XML format through Appium and transforms it into HTML format through a heuristic algorithm. 
Subsequently, as illustrated in Figure \ref{in_model} leveraging the HTML, task details, and APP list, LLM generates a comprehensive task plan, outlining the necessary applications and corresponding sub-tasks. 
As the collection of APIs is organized based on the classification of APPs, we can get the API set that may be used in plan. 

The task plan is executed iteratively. In each iteration, the model either performs an API call or a UI operation. After each execution, the model records the success or failure of the action in its history, generates the subsequent thought, and evaluates whether the task has been completed. For the actual running process of an algorithm, please refer to the appendix \ref{Algorithm Examples}.

\begin{algorithm}[t]
\small
\caption{Baseline Model}

\begin{algorithmic}[1]
\Require
    description of the Task, $Task$;
    APP list, $L_{APP}$;
    API list, $L_{API}$;
    max loop step, $M_\text{step}$;
    initial thought, $Tho$;
\Ensure
    actions history, $A\!H$;
    total steps, $Step$;
    finish flag, $Finish$;
\State $Html \gets Appium(Emulator)$
\State $Plan \gets LLM(Task, L_{APP})$
\State $Step=0$, \ $Finish = False$
\State $A\!H=[ \ ]$ \ 

\While {$(Step \leq M_{step})  and  (Finish \neq True)$} 
    \State $Step++$;
    \State $Html \gets Appium(Emulator)$;
    
    \State $API \!\! \gets \!\! LLM(Task, L_{API}, A\!H, Tho, Plan, Html)$
    
    \If {$ API$}
        \State $Action(API)$
        \State $A\!H.append(API)$
    \Else
        \State $UI \gets LLM(Task, A\!H, Tho, Plan, Html)$
        \State $Action(UI)$
        \State $A\!H.append(UI)$
    \EndIf
    \State $Html \gets Appium(Emulator)$;
    \State $Tho \gets LLM(Task, A\!H, Plan, Html)$
    \State $Finish \gets LLM(Task, A\!H, Tho,  Html)$

\EndWhile
\end{algorithmic} 
\label{alg:example} 
\end{algorithm}

\begin{table*}[t]
\centering
\renewcommand{\arraystretch}{1.2}
\begin{adjustbox}{max width=\textwidth}
\footnotesize
\begin{tabular}{@{}lcccccccccccc@{}}
    \toprule[1pt]
    \multirow{2}{*}{\textbf{Metric}} & \multicolumn{3}{c}{LLaMA-13B} & \multicolumn{3}{c}{LLaMA-70B} & \multicolumn{3}{c}{GPT-3.5-turbo} & \multicolumn{3}{c}{GPT-4} \\
    \cmidrule(lr){2-4} \cmidrule(lr){5-7} \cmidrule(lr){8-10} \cmidrule(lr){11-13}
    & {\footnotesize SAST} & {\footnotesize SAMT} & {\footnotesize MAMT} & {\footnotesize SAST} & {\footnotesize SAMT} & {\footnotesize MAMT} & {\footnotesize SAST} & {\footnotesize SAMT} & {\footnotesize MAMT} & {\footnotesize SAST} & {\footnotesize SAMT} & {\footnotesize MAMT} \\
\hline
\textbf{Average \#Steps} & 7.43 & 18.76 & 49.52 & 5.97 & 16.63 & 48.91 & 4.53 & \textbf{12.06} & 48.73 & \textbf{3.79} & 13.94 & \textbf{44.86}  \\
\textbf{PassRate} & 44.58 & 27.67 & 8 & 56.62 & 54 & 13.5 & 64.94 & \textbf{64} & 15.5 & \textbf{80.96} & 63 &  \textbf{26.5} \\
\textbf{CheckPoint\textsubscript{l1}} & 46.08 & 43.67 & 28.74 & 56.62 & 61 & 39.98 & 66.75 & 67 & 43.16 & \textbf{81.57} & \textbf{72.66} & \textbf{61.34} \\
\textbf{CheckPoint\textsubscript{l2}} & 34.85 & 29.13 & 21.39 & 63.12 & 62.73 & 41.21 & 76.21 & 71.29 & 44.09 & \textbf{83.76} & \textbf{77.35} & \textbf{52.98} \\
\hline
\end{tabular}
\end{adjustbox}
\caption{Results of the agents based on different LLMs on Mobile-Bench dataset. On MAMT data, due to context length limitations, a compression is applied to the actions history by retaining only the most recent 20 entries.}
\label{tab:mainresult}
\end{table*}

\subsection{Setup}
We evaluate four popular LLMs on the proposed Mobile-Bench task set: GPT-3.5-turbo \cite{ouyang2022training}, GPT-4 \cite{nori2023capabilities},  LLaMA-13B  and LLaMA-70B\cite{touvron2023llama}, while ChatGPT-3.5 and GPT-4 are accessed through the online APIs of OpenAI. The experiments are conducted with a 3-shot in-context learning under sampling temperature of 0.1. 
Recognizing that task execution incurs costs, we preset different maximum step limits for tasks based on their difficulty levels. For the three categories of SAST, SAMT, and MAMT, we set the max step to 10, 20, and 50 respectively.
Owing to the limit of budget, only GPT-3.5 utilizes an interface with a context length of 16K. GPT-4 uses a standard interface, which necessitated compression and trimming of actions history. 
 See Appendix \ref{setting} for other settings. 
\subsection{Results}
As observed in Table \ref{tab:mainresult}, it can be observed that GPT-3.5 outperforms GPT-4 in PassRate on SAMT(64\%>63\%), and it requires fewer steps to complete the task(12.06<13.94). To investigate this phenomenon, we analyze the output files and find that models with poorer performance exhibit PassRate misjudgments: they prematurely terminate even when the task is not completed. This phenomenon is also present in LLaMA, which exhibits a high PassRate (44.58\%) but low CheckPoint coverage (34.85\%). At the same time, we delved into why the results for MAMT are so low(15.5\%, 26.5\%). Our analysis revealed that LLMs often exhibit greedy exploration behavior when completing tasks, meaning they struggle to determine when to exit the current application and transition to the next one. This tendency is particularly prevalent in certain generation tasks.  Moreover, as the actions history increases, its ability to accurately judge task progress becomes increasingly challenging. For more detailed result, please refer to Table \ref{tab:app_category}.
\begin{table}[htbp]
\centering
\renewcommand{\arraystretch}{1.2}
\begin{adjustbox}{max width=\textwidth}
\resizebox{\columnwidth}{!}{
\begin{tabular}{@{}lccc@{}}
\hline
\textbf{Settings} & \textbf{Average \#Steps} & \textbf{CheckPoint\textsubscript{l2}} & \textbf{PassRate}\\
 \hline
 SAST (GPT-4) & \textbf{3.79} & \textbf{83.76} & \textbf{80.96}  \\
 SAMT (GPT-4) & 13.94 & 77.35 & 63 \\
 MAMT (GPT-4) & 44.86 & 52.98 & 26.5 \\
 \hline
 SAST (w/o API) & 6.13 & 72.73 & 74.39 \\
 SAMT (w/o API) & 16.86 & 56.74 & 48 \\
 MAMT (w/o API) & 49.17 & 31.69 & 9.5 \\
\hline
\end{tabular}
}
\end{adjustbox}
\caption{API Ablation Study based on GPT-4.}
\label{tab: APIcheck}
\end{table}

\subsection{Impact of API Calls}
API Calls can accelerate task execution, as a single call often replaces several sequential UI steps.  From another perspective, the ability of the agent to select appropriate APIs and input parameters warrants further investigation. Choosing the wrong API may lead the task in an incorrect direction or require a significant number of steps to rectify. Therefore, in Table \ref{tab: APIcheck}, we evaluate and analyze the impact of introducing APIs on task completion based on GPT-4. 

From Table \ref{tab: APIcheck}, it can be seen that even in SAST, the PassRate has decreased by 6.57\% (from 80.96 to 74.39). Furthermore, the values for CheckPoints\textsubscript{l2} exhibit a more pronounced decrease after API removal, with a drop exceeding 20\% in SAMT. Simultaneously, we have observed varying increases in the average number of steps, which align with our expectations. 
We analyzed the results and found that the inability to accurately scroll pages, inefficient exploration of page functionality, and failure to click graphical buttons are the primary reasons for the low efficiency of UI operations.

\begin{table}[htbp]
\centering
\renewcommand{\arraystretch}{1.3}
\Large 
\begin{adjustbox}{max width=0.48\textwidth}
\begin{tabular}{@{}lcccc@{}}
\hline
\textbf{Settings} & \textbf{Average \#Steps} & \textbf{CheckPoint\textsubscript{l1}} & \textbf{CheckPoint\textsubscript{l2}} & \textbf{PassRate}\\
\hline
SAST (GPT-4) & \textbf{3.63} & \textbf{82} & \textbf{79.74} & \textbf{76} \\
SAST (w/o thought) & 8.86 & \textbf{82} & 29.16 & 24\\
SAST (w/o plan) & 3.98 & 76 & 74.54 & 72\\
\hline
SAMT (GPT-4) & \textbf{13.94} & \textbf{63} & \textbf{72.66} & \textbf{77} \\
SAMT (w/o thought) & 19.54 & \textbf{63} & 18.31 & 20\\
SAMT (w/o plan) & 17.09 & 52 & 58.02 & 62 \\
\hline
\end{tabular}
\end{adjustbox}
\caption{Thought and Plan Ablation Study on SAST (subset 50) and SAMT (subset 200) based on GPT-4.}
\label{tab:thoughtcheck}
\end{table}

\subsection{Impact of Plan and Thought}
Since observation-thought-action is already a standardized process in the agent direction\cite{qin2023toolllm}, and verified by experimental results, planning and thought before action are essential. 
From the experimental results, we can find that without the observation-thought step, the agent is almost unable to complete the task(77->20, 76->24), which is because it cannot determine the next action category and the current task status. In more complex tasks SAMT, losing the plan has more negative consequences(77->62). But they will have almost no impact on \textit{CheckPoint\textsubscript{l1}}(82->82 63->63), because the application selection is almost done by the API Call.

\section{Conclusion}
In this work we have proposed an agent capability testing environment that supports API and UI interaction on mobile phone. This holds significant importance for exploring how LLMs can be integrated with mobile operating systems. Additionally, it can serve as a valuable reference for developing testing platforms for operating systems to evaluate the capabilities of LLM agents. 
We collected and released a test dataset containing tasks for multiple APPs, ensuring its quality through human verification. 
Based on this data set and environment, we tested the planning, decision-making and execution of various LLM-based agents. Please refer to the Section \ref{limitation} for the limitations of our benchmark.

\section{Limitations}
\label{limitation}
While general large models exhibit strong capabilities in reasoning and planning, they tend to have pronounced illusions in API calls. As a result, the language model may become confused about the application's functionality, leading to a reluctance to continue and complete the task. Therefore, fine-tuning a model for instructions is highly necessary. 

Automatic CheckPoint is a process evaluation metric, making it challenging to assess the quality of the final outcome. This depends on whether the agent has obtained the necessary information (actions) on the required pages. 

The enhancement of the agent's capabilities relies on extensive API and SDK libraries, requiring substantial support from application development companies. 

\section{Ethics Statement}
We have rigorously refined our dataset to remove any elements that could compromise personal privacy, thereby guaranteeing the highest level of protection for individual data. The evaluation of our work was carried out through a meticulously randomized selection of IT professionals. This process ensured a gender-balanced and educationally diverse panel, reflecting a wide spectrum of perspectives and expertise.

\section{Acknowledgements}
We thank the Xiaoai Voice Department of Xiaomi Technology Corporation for their raw data support for this project. We additionally thank our crowd annotators for their diligent work, Junfeng Peng and Yifan Cheng for contributing to the human performance estimates, and the anonymous reviewers for their constructive comments. 
This work was supported by the NSFC (U2001212, 62032001, and 61932004). 


\bibliography{mobile-bench}

\begin{thebibliography}{48}
\expandafter\ifx\csname natexlab\endcsname\relax\def\natexlab#1{#1}\fi

\bibitem[{Abdolmaleki et~al.(2018)Abdolmaleki, Springenberg, Tassa, Munos, Heess, and Riedmiller}]{abdolmaleki2018maximum}
Abbas Abdolmaleki, Jost~Tobias Springenberg, Yuval Tassa, Remi Munos, Nicolas Heess, and Martin Riedmiller. 2018.
\newblock Maximum a posteriori policy optimisation.
\newblock \emph{arXiv preprint arXiv:1806.06920}.

\bibitem[{Bai et~al.(2022)Bai, Kadavath, Kundu, Askell, Kernion, Jones, Chen, Goldie, Mirhoseini, McKinnon et~al.}]{bai2022constitutional}
Y~Bai, S~Kadavath, S~Kundu, A~Askell, J~Kernion, A~Jones, A~Chen, A~Goldie, A~Mirhoseini, C~McKinnon, et~al. 2022.
\newblock Constitutional ai: Harmlessness from ai feedback (arxiv: 2212.08073). arxiv.

\bibitem[{Bajunaid and Menasc{\'e}(2018)}]{bajunaid2018efficient}
Noor Bajunaid and Daniel~A Menasc{\'e}. 2018.
\newblock Efficient modeling and optimizing of checkpointing in concurrent component-based software systems.
\newblock \emph{Journal of Systems and Software}, 139:1--13.

\bibitem[{Barth-Maron et~al.(2018)Barth-Maron, Hoffman, Budden, Dabney, Horgan, Tb, Muldal, Heess, and Lillicrap}]{barth2018distributed}
Gabriel Barth-Maron, Matthew~W Hoffman, David Budden, Will Dabney, Dan Horgan, Dhruva Tb, Alistair Muldal, Nicolas Heess, and Timothy Lillicrap. 2018.
\newblock Distributed distributional deterministic policy gradients.
\newblock \emph{arXiv preprint arXiv:1804.08617}.

\bibitem[{Bolotova-Baranova et~al.(2023)Bolotova-Baranova, Blinov, Filippova, Scholer, and Sanderson}]{bolotova2023wikihowqa}
Valeriia Bolotova-Baranova, Vladislav Blinov, Sofya Filippova, Falk Scholer, and Mark Sanderson. 2023.
\newblock Wikihowqa: A comprehensive benchmark for multi-document non-factoid question answering.
\newblock In \emph{Proceedings of the 61st Annual Meeting of the Association for Computational Linguistics (Volume 1: Long Papers)}, pages 5291--5314.

\bibitem[{Bolt(1980)}]{bolt1980put}
Richard~A Bolt. 1980.
\newblock “put-that-there” voice and gesture at the graphics interface.
\newblock In \emph{Proceedings of the 7th annual conference on Computer graphics and interactive techniques}, pages 262--270.

\bibitem[{Brown et~al.(2020)Brown, Mann, Ryder, Subbiah, Kaplan, Dhariwal, Neelakantan, Shyam, Sastry, Askell et~al.}]{brown2020language}
Tom Brown, Benjamin Mann, Nick Ryder, Melanie Subbiah, Jared~D Kaplan, Prafulla Dhariwal, Arvind Neelakantan, Pranav Shyam, Girish Sastry, Amanda Askell, et~al. 2020.
\newblock Language models are few-shot learners.
\newblock \emph{Advances in neural information processing systems}, 33:1877--1901.

\bibitem[{Bubeck et~al.(2023)Bubeck, Chandrasekaran, Eldan, Gehrke, Horvitz, Kamar, Lee, Lee, Li, Lundberg et~al.}]{bubeck2023sparks}
S{\'e}bastien Bubeck, Varun Chandrasekaran, Ronen Eldan, Johannes Gehrke, Eric Horvitz, Ece Kamar, Peter Lee, Yin~Tat Lee, Yuanzhi Li, Scott Lundberg, et~al. 2023.
\newblock Sparks of artificial general intelligence: Early experiments with gpt-4.
\newblock \emph{arXiv preprint arXiv:2303.12712}.

\bibitem[{Chowdhery et~al.(2022)Chowdhery, Narang, Devlin, Bosma, Mishra, Roberts, Barham, Chung, Sutton, Gehrmann et~al.}]{chowdhery2022palm}
Aakanksha Chowdhery, Sharan Narang, Jacob Devlin, Maarten Bosma, Gaurav Mishra, Adam Roberts, Paul Barham, Hyung~Won Chung, Charles Sutton, Sebastian Gehrmann, et~al. 2022.
\newblock Palm: Scaling language modeling with pathways.
\newblock \emph{arXiv preprint arXiv:2204.02311}.

\bibitem[{Chowdhery et~al.(2023)Chowdhery, Narang, Devlin, Bosma, Mishra, Roberts, Barham, Chung, Sutton, Gehrmann et~al.}]{chowdhery2023palm}
Aakanksha Chowdhery, Sharan Narang, Jacob Devlin, Maarten Bosma, Gaurav Mishra, Adam Roberts, Paul Barham, Hyung~Won Chung, Charles Sutton, Sebastian Gehrmann, et~al. 2023.
\newblock Palm: Scaling language modeling with pathways.
\newblock \emph{Journal of Machine Learning Research}, 24(240):1--113.

\bibitem[{Deka et~al.(2017)Deka, Huang, Franzen, Hibschman, Afergan, Li, Nichols, and Kumar}]{deka2017rico}
Biplab Deka, Zifeng Huang, Chad Franzen, Joshua Hibschman, Daniel Afergan, Yang Li, Jeffrey Nichols, and Ranjitha Kumar. 2017.
\newblock Rico: A mobile app dataset for building data-driven design applications.
\newblock In \emph{Proceedings of the 30th annual ACM symposium on user interface software and technology}, pages 845--854.

\bibitem[{Desnos and Gueguen(2011)}]{desnos2011android}
Anthony Desnos and Geoffroy Gueguen. 2011.
\newblock Android: From reversing to decompilation.
\newblock \emph{Proc. of Black Hat Abu Dhabi}, 1:1--24.

\bibitem[{Du et~al.(2021)Du, Qian, Liu, Ding, Qiu, Yang, and Tang}]{du2021glm}
Zhengxiao Du, Yujie Qian, Xiao Liu, Ming Ding, Jiezhong Qiu, Zhilin Yang, and Jie Tang. 2021.
\newblock Glm: General language model pretraining with autoregressive blank infilling.
\newblock \emph{arXiv preprint arXiv:2103.10360}.

\bibitem[{Espeholt et~al.(2018)Espeholt, Soyer, Munos, Simonyan, Mnih, Ward, Doron, Firoiu, Harley, Dunning et~al.}]{espeholt2018impala}
Lasse Espeholt, Hubert Soyer, Remi Munos, Karen Simonyan, Vlad Mnih, Tom Ward, Yotam Doron, Vlad Firoiu, Tim Harley, Iain Dunning, et~al. 2018.
\newblock Impala: Scalable distributed deep-rl with importance weighted actor-learner architectures.
\newblock In \emph{International conference on machine learning}, pages 1407--1416. PMLR.

\bibitem[{Fan et~al.(2024)Fan, Jiang, Li, Meng, Han, Shang, Sun, Wang, and Wang}]{fan2024not}
Siqi Fan, Xin Jiang, Xiang Li, Xuying Meng, Peng Han, Shuo Shang, Aixin Sun, Yequan Wang, and Zhongyuan Wang. 2024.
\newblock Not all layers of llms are necessary during inference.
\newblock \emph{arXiv preprint arXiv:2403.02181}.

\bibitem[{F{\o}lstad and Brandtz{\ae}g(2017)}]{folstad2017chatbots}
Asbj{\o}rn F{\o}lstad and Petter~Bae Brandtz{\ae}g. 2017.
\newblock Chatbots and the new world of hci.
\newblock \emph{interactions}, 24(4):38--42.

\bibitem[{Guo et~al.(2023)Guo, Zhang, Liang, Zhao, and Nan}]{guo2023pptc}
Yiduo Guo, Zekai Zhang, Yaobo Liang, Dongyan Zhao, and Duan Nan. 2023.
\newblock Pptc benchmark: Evaluating large language models for powerpoint task completion.
\newblock \emph{arXiv preprint arXiv:2311.01767}.

\bibitem[{Kapturowski et~al.(2018)Kapturowski, Ostrovski, Quan, Munos, and Dabney}]{kapturowski2018recurrent}
Steven Kapturowski, Georg Ostrovski, John Quan, Remi Munos, and Will Dabney. 2018.
\newblock Recurrent experience replay in distributed reinforcement learning.
\newblock In \emph{International conference on learning representations}.

\bibitem[{Karat et~al.(2002)Karat, Vergo, and Nahamoo}]{karat2002conversational}
Clare-Marie Karat, John Vergo, and David Nahamoo. 2002.
\newblock Conversational interface technologies.
\newblock \emph{The human-computer interaction handbook}, pages 169--186.

\bibitem[{Li et~al.(2021)Li, Popowski, Mitchell, and Myers}]{li2021screen2vec}
Toby Jia-Jun Li, Lindsay Popowski, Tom Mitchell, and Brad~A Myers. 2021.
\newblock Screen2vec: Semantic embedding of gui screens and gui components.
\newblock In \emph{Proceedings of the 2021 CHI Conference on Human Factors in Computing Systems}, pages 1--15.

\bibitem[{Li et~al.(2020)Li, He, Zhou, Zhang, and Baldridge}]{li2020mapping}
Yang Li, Jiacong He, Xin Zhou, Yuan Zhang, and Jason Baldridge. 2020.
\newblock Mapping natural language instructions to mobile ui action sequences.
\newblock \emph{arXiv preprint arXiv:2005.03776}.

\bibitem[{Liu et~al.(2023{\natexlab{a}})Liu, Yu, Zhang, Xu, Lei, Lai, Gu, Ding, Men, Yang et~al.}]{liu2023agentbench}
Xiao Liu, Hao Yu, Hanchen Zhang, Yifan Xu, Xuanyu Lei, Hanyu Lai, Yu~Gu, Hangliang Ding, Kaiwen Men, Kejuan Yang, et~al. 2023{\natexlab{a}}.
\newblock Agentbench: Evaluating llms as agents.
\newblock \emph{arXiv preprint arXiv:2308.03688}.

\bibitem[{Liu et~al.(2019)Liu, Ott, Goyal, Du, Joshi, Chen, Levy, Lewis, Zettlemoyer, and Stoyanov}]{liu2019roberta}
Yinhan Liu, Myle Ott, Naman Goyal, Jingfei Du, Mandar Joshi, Danqi Chen, Omer Levy, Mike Lewis, Luke Zettlemoyer, and Veselin Stoyanov. 2019.
\newblock Roberta: A robustly optimized bert pretraining approach.
\newblock \emph{arXiv preprint arXiv:1907.11692}.

\bibitem[{Liu et~al.(2024)Liu, Chen, Zhang, Gao, Zhang, and Yan}]{liu2024skepticism}
Yuhan Liu, Xiuying Chen, Xiaoqing Zhang, Xing Gao, Ji~Zhang, and Rui Yan. 2024.
\newblock From skepticism to acceptance: Simulating the attitude dynamics toward fake news.
\newblock \emph{arXiv preprint arXiv:2403.09498}.

\bibitem[{Liu et~al.(2023{\natexlab{b}})Liu, Chen, Wang, Chen, Wu, Che, Wang, and Wang}]{liu2023chatting}
Zhe Liu, Chunyang Chen, Junjie Wang, Mengzhuo Chen, Boyu Wu, Xing Che, Dandan Wang, and Qing Wang. 2023{\natexlab{b}}.
\newblock Chatting with gpt-3 for zero-shot human-like mobile automated gui testing.
\newblock \emph{arXiv preprint arXiv:2305.09434}.

\bibitem[{Mnih et~al.(2015)Mnih, Kavukcuoglu, Silver, Rusu, Veness, Bellemare, Graves, Riedmiller, Fidjeland, Ostrovski et~al.}]{mnih2015human}
Volodymyr Mnih, Koray Kavukcuoglu, David Silver, Andrei~A Rusu, Joel Veness, Marc~G Bellemare, Alex Graves, Martin Riedmiller, Andreas~K Fidjeland, Georg Ostrovski, et~al. 2015.
\newblock Human-level control through deep reinforcement learning.
\newblock \emph{nature}, 518(7540):529--533.

\bibitem[{Nori et~al.(2023)Nori, King, McKinney, Carignan, and Horvitz}]{nori2023capabilities}
Harsha Nori, Nicholas King, Scott~Mayer McKinney, Dean Carignan, and Eric Horvitz. 2023.
\newblock Capabilities of gpt-4 on medical challenge problems.
\newblock \emph{arXiv preprint arXiv:2303.13375}.

\bibitem[{OpenAI(2023)}]{openai2023gpt4}
OpenAI. 2023.
\newblock \href {http://arxiv.org/abs/2303.08774} {Gpt-4 technical report}.

\bibitem[{Ouyang et~al.(2022)Ouyang, Wu, Jiang, Almeida, Wainwright, Mishkin, Zhang, Agarwal, Slama, Ray et~al.}]{ouyang2022training}
Long Ouyang, Jeffrey Wu, Xu~Jiang, Diogo Almeida, Carroll Wainwright, Pamela Mishkin, Chong Zhang, Sandhini Agarwal, Katarina Slama, Alex Ray, et~al. 2022.
\newblock Training language models to follow instructions with human feedback.
\newblock \emph{Advances in Neural Information Processing Systems}, 35:27730--27744.

\bibitem[{Qin et~al.(2023)Qin, Liang, Ye, Zhu, Yan, Lu, Lin, Cong, Tang, Qian et~al.}]{qin2023toolllm}
Yujia Qin, Shihao Liang, Yining Ye, Kunlun Zhu, Lan Yan, Yaxi Lu, Yankai Lin, Xin Cong, Xiangru Tang, Bill Qian, et~al. 2023.
\newblock Toolllm: Facilitating large language models to master 16000+ real-world apis.
\newblock \emph{arXiv preprint arXiv:2307.16789}.

\bibitem[{Rawles et~al.(2023)Rawles, Li, Rodriguez, Riva, and Lillicrap}]{rawles2023android}
Christopher Rawles, Alice Li, Daniel Rodriguez, Oriana Riva, and Timothy Lillicrap. 2023.
\newblock Android in the wild: A large-scale dataset for android device control.
\newblock \emph{arXiv preprint arXiv:2307.10088}.

\bibitem[{Schick et~al.(2023)Schick, Dwivedi-Yu, Dess{\`\i}, Raileanu, Lomeli, Zettlemoyer, Cancedda, and Scialom}]{schick2023toolformer}
Timo Schick, Jane Dwivedi-Yu, Roberto Dess{\`\i}, Roberta Raileanu, Maria Lomeli, Luke Zettlemoyer, Nicola Cancedda, and Thomas Scialom. 2023.
\newblock Toolformer: Language models can teach themselves to use tools.
\newblock \emph{arXiv preprint arXiv:2302.04761}.

\bibitem[{Sengupta et~al.(2023)Sengupta, Singh, and Vinjit}]{sengupta2023platform}
Aritro Sengupta, Amit Singh, and BM~Vinjit. 2023.
\newblock A platform independent and forensically sound method to extract whatsapp data from mobile phones.
\newblock \emph{International Journal of Electronic Security and Digital Forensics}, 15(3):259--280.

\bibitem[{Shen et~al.(2023)Shen, Song, Tan, Li, Lu, and Zhuang}]{shen2023hugginggpt}
Yongliang Shen, Kaitao Song, Xu~Tan, Dongsheng Li, Weiming Lu, and Yueting Zhuang. 2023.
\newblock Hugginggpt: Solving ai tasks with chatgpt and its friends in huggingface.
\newblock \emph{arXiv preprint arXiv:2303.17580}.

\bibitem[{Shi et~al.(2017)Shi, Karpathy, Fan, Hernandez, and Liang}]{shi2017world}
Tianlin Shi, Andrej Karpathy, Linxi Fan, Jonathan Hernandez, and Percy Liang. 2017.
\newblock World of bits: An open-domain platform for web-based agents.
\newblock In \emph{International Conference on Machine Learning}, pages 3135--3144. PMLR.

\bibitem[{Sun et~al.(2024{\natexlab{a}})Sun, Lin, Yan, Zhu, Song, Gao, Shang, and Yan}]{sun2024facilitating}
Hongda Sun, Hongzhan Lin, Haiyu Yan, Chen Zhu, Yang Song, Xin Gao, Shuo Shang, and Rui Yan. 2024{\natexlab{a}}.
\newblock Facilitating multi-role and multi-behavior collaboration of large language models for online job seeking and recruiting.
\newblock \emph{arXiv preprint arXiv:2405.18113}.

\bibitem[{Sun et~al.(2024{\natexlab{b}})Sun, Liu, Wu, Yan, Tai, Gao, Shang, and Yan}]{sun2024harnessing}
Hongda Sun, Yuxuan Liu, Chengwei Wu, Haiyu Yan, Cheng Tai, Xin Gao, Shuo Shang, and Rui Yan. 2024{\natexlab{b}}.
\newblock Harnessing multi-role capabilities of large language models for open-domain question answering.
\newblock In \emph{Proceedings of the ACM on Web Conference 2024}, pages 4372--4382.

\bibitem[{Sun et~al.(2023)Sun, Xu, Liu, Luan, Wang, Shang, Wen, and Yan}]{sun2023determlr}
Hongda Sun, Weikai Xu, Wei Liu, Jian Luan, Bin Wang, Shuo Shang, Ji-Rong Wen, and Rui Yan. 2023.
\newblock Determlr: Augmenting llm-based logical reasoning from indeterminacy to determinacy.
\newblock \emph{arXiv preprint arXiv:2310.18659}.

\bibitem[{Touvron et~al.(2023)Touvron, Lavril, Izacard, Martinet, Lachaux, Lacroix, Rozi{\`e}re, Goyal, Hambro, Azhar et~al.}]{touvron2023llama}
Hugo Touvron, Thibaut Lavril, Gautier Izacard, Xavier Martinet, Marie-Anne Lachaux, Timoth{\'e}e Lacroix, Baptiste Rozi{\`e}re, Naman Goyal, Eric Hambro, Faisal Azhar, et~al. 2023.
\newblock Llama: Open and efficient foundation language models.
\newblock \emph{arXiv preprint arXiv:2302.13971}.

\bibitem[{Toyama et~al.(2021)Toyama, Hamel, Gergely, Comanici, Glaese, Ahmed, Jackson, Mourad, and Precup}]{toyama2021androidenv}
Daniel Toyama, Philippe Hamel, Anita Gergely, Gheorghe Comanici, Amelia Glaese, Zafarali Ahmed, Tyler Jackson, Shibl Mourad, and Doina Precup. 2021.
\newblock Androidenv: A reinforcement learning platform for android.
\newblock \emph{arXiv preprint arXiv:2105.13231}.

\bibitem[{Wang et~al.(2023)Wang, Li, and Li}]{wang2023enabling}
Bryan Wang, Gang Li, and Yang Li. 2023.
\newblock Enabling conversational interaction with mobile ui using large language models.
\newblock In \emph{Proceedings of the 2023 CHI Conference on Human Factors in Computing Systems}, pages 1--17.

\bibitem[{Wang et~al.(2021)Wang, Li, Zhou, Chen, Grossman, and Li}]{wang2021screen2words}
Bryan Wang, Gang Li, Xin Zhou, Zhourong Chen, Tovi Grossman, and Yang Li. 2021.
\newblock Screen2words: Automatic mobile ui summarization with multimodal learning.
\newblock In \emph{The 34th Annual ACM Symposium on User Interface Software and Technology}, pages 498--510.

\bibitem[{Wen et~al.(2023{\natexlab{a}})Wen, Li, Liu, Zhao, Yu, Li, Jiang, Liu, Zhang, and Liu}]{wen2023empowering}
Hao Wen, Yuanchun Li, Guohong Liu, Shanhui Zhao, Tao Yu, Toby Jia-Jun Li, Shiqi Jiang, Yunhao Liu, Yaqin Zhang, and Yunxin Liu. 2023{\natexlab{a}}.
\newblock Empowering llm to use smartphone for intelligent task automation.
\newblock \emph{arXiv preprint arXiv:2308.15272}.

\bibitem[{Wen et~al.(2023{\natexlab{b}})Wen, Wang, Liu, and Li}]{wen2023droidbot}
Hao Wen, Hongming Wang, Jiaxuan Liu, and Yuanchun Li. 2023{\natexlab{b}}.
\newblock Droidbot-gpt: Gpt-powered ui automation for android.
\newblock \emph{arXiv preprint arXiv:2304.07061}.

\bibitem[{Yao et~al.(2022)Yao, Chen, Yang, and Narasimhan}]{yao2022webshop}
Shunyu Yao, Howard Chen, John Yang, and Karthik Narasimhan. 2022.
\newblock Webshop: Towards scalable real-world web interaction with grounded language agents.
\newblock \emph{Advances in Neural Information Processing Systems}, 35:20744--20757.

\bibitem[{Zhang and Van~Huynh(2023)}]{zhang2023deep}
Bolun Zhang and Nguyen Van~Huynh. 2023.
\newblock Deep deterministic policy gradient for end-to-end communication systems without prior channel knowledge.
\newblock \emph{arXiv preprint arXiv:2305.07448}.

\bibitem[{Zhang et~al.(2023)Zhang, Chen, and Yu}]{zhang2023mobile}
Danyang Zhang, Lu~Chen, and Kai Yu. 2023.
\newblock Mobile-env: A universal platform for training and evaluation of mobile interaction.
\newblock \emph{arXiv preprint arXiv:2305.08144}.

\bibitem[{Zhang et~al.(2022)Zhang, Roller, Goyal, Artetxe, Chen, Chen, Dewan, Diab, Li, Lin et~al.}]{zhang2022opt}
Susan Zhang, Stephen Roller, Naman Goyal, Mikel Artetxe, Moya Chen, Shuohui Chen, Christopher Dewan, Mona Diab, Xian Li, Xi~Victoria Lin, et~al. 2022.
\newblock Opt: Open pre-trained transformer language models.
\newblock \emph{arXiv preprint arXiv:2205.01068}.

\end{thebibliography}

\clearpage
\appendix

\onecolumn

\section{Settings}
\label{setting}
We conduct experiments on the Android 14.0 version emulator and use Appium UiAutomator2 Driver for automated testing.  Before each execution of a task, we load a snapshot to ensure the emulator in the same environment every time. For all applications, we have logged in to the account in advance to ensure that the full function of the application can be used. Since we tests in the real world, we filtered out any tasks that included payments.

\section{Details of Dataset}
\subsection{Dataset quality analysis}
\label{Dataset quality analysis}
The root cause of low-quality data often lies in the inaccuracies in the descriptions of applications. Additionally, ambiguity in query generation also plays a significant role. For example, in the query \textit{''Help me find pictures related to Beijing''}, although the user has not explicitly specified the source application, for a human, the expected result would likely be a search engine or a map application, as the images are not likely to be from the user themselves. However, for LLM, because the statement includes the word ``pictures'', it might be reasonable for it to spend all its time searching for pictures in the gallery application, even though this effort would ultimately be in vain. CheckPoint coverage is calculated as the weighted sum of the scores for the three types of CheckPoints mentioned above.
\subsection{Prompts for Instruction Generation}
Below we list the detailed prompt for instruction generation, including single-APP-multi-task description, multi-APP-multi-task description.

\textbf{single-APP-multi-task description:} 

You will be provided with an application with descriptions, an available API list including adb command, function description and parameter information. You should create 5 varied, innovative, and detailed multi task queries that employ this application as a tool, API can be used as an auxiliary.  

Each query should include the necessary parameters. Note that you shouldn’t ask ‘which APP to use’, rather, simply state your needs that can be addressed by these APPs. You should also avoid asking for the input parameters required by the APP call, but instead directly provide the parameter in your query. Those related APP and APIs have to strictly come from the provided lists. 

At the same time, you also need to provide the CheckPoint of this query, including package, key phrase and API. The package comes from the package corresponding to the APP to be used. Key phrase is the key click element or key input character that the Android emulator will perform when executing this query, which is used to check whether the query has been completed. Key phrase should be noun and part of query, should be kept as short as possible. 

Key phrase can contain multiple pieces of information, "$|$" means the query passes when any of the following texts are completed. "$|$" is used to separate synonymous expressions of the same noun; "$ \& $" indicates that the query must be passed when all texts are completed; sequential CheckPoints are stored in "$[ \ ]$", and the count increases by one for each passed element. The "ADB Command" to be used is stored in the API, which may also be empty.

Deliver your response in this format: \\
$[ \{$ \\
\indent"id": "number" \\
\indent"query": "text"\\
\indent"APP": "APP name"\\
\indent"CheckPoint": $\{$\\
\indent\indent "package": "APP package name" \\
\indent\indent"key phrase": $[$"text1", ... $]$ \\
\indent\indent"API: $[$"API1", ... $ ]$"       \\
\indent\indent$\}$ \\
$\ \}$ \\
... \\
$]$ 

\textbf{multi-APP-multi-task description:} 

You will be provided with some APPs with descriptions, available API list including adb command, function description and parameter information. You should create 3 varied, innovative, and detailed multi queries that employ multi-APP as a tool, API can be used as an auxiliary. 

Each query should include the necessary parameters. Note that you shouldn’t ask ‘which APP to use’, rather, simply state your needs that can be addressed by these APPs. You should also avoid asking for the input parameters required by the APP call, but instead directly provide the parameter in your query. Those related APPs and APIs have to strictly come from the provided lists. You should first think about possible related APP combinations, then give your query. Keep in mind that each query should call upon two to four APPs. 

At the same time, you also need to provide the CheckPoint of this query, including package, key phrase and API. The package comes from the package corresponding to the APP to be used. Key phrase is the key click element or key input character that the Android emulator will perform when executing this query, which is used to check whether the query has been completed. Key phrase should be noun and part of query, should be kept as short as possible. 

Key phrase can contain multiple pieces of information, "$|$" means the query passes when any of the following texts are completed. "$|$" is used to separate synonymous expressions of the same noun; "$\&$" indicates that the query must be passed when all texts are completed; sequential CheckPoints are stored in "$[ \ ]$", and the count increases by one for each passed element. The "ADB Command" to be used is stored in the API, which may also be empty. For different queries, overlap of related APPs should be as little as possible.

Deliver your response in this format: \\
$[ \{$ \\
\indent"id": "number" \\
\indent"query": "text"\\
\indent"APP": $[$"APP name1", ... \ $]$ \\
\indent"CheckPoint": $\{$\\
\indent\indent "package": $[$"APP package name1", ... \ $]$ \\
\indent\indent"key phrase": $[$"text1", ... $]$ \\
\indent\indent"API: $[$"API1", ... $ ]$"       \\
\indent\indent$\}$ \\
$\ \}$ \\
... \\
$]$

\subsection{APP\&API statistics}
\label{APP&API statistics}

\begin{figure*}[th]
  \centering
        \includegraphics[width=0.8\textwidth, page=1]{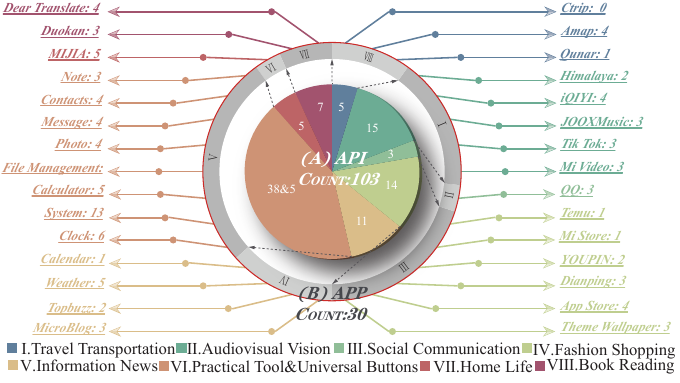}
        \caption{APP classification and quantity chart: The largest category is utility tools, where we categorize fundamental mobile applications. Their distinctive feature is the use of standard API interfaces, and the API functionality is more comprehensive.}
        \label{fig:image1_APPAPI}
\end{figure*}
As can be seen from Figure \ref{fig:image1_APPAPI}, each functional area contains at least one application and its corresponding API. These applications are sufficient to meet the daily needs of users. In other words, our simulation environment is almost consistent with the real daily use environment, and it is consistent with the real daily use environment. Open world information exchange. There are so many practical tools that are the basic functions of mobile . They have been automatically installed and completed during system installation, and standard API interfaces for tools are easier to obtain. Our next step is to increase the number of APIs and SDKs for third-party applications.

\subsection{Case study}
\label{case_APPendix}
CheckPoints is a group of words, including packages, key phases, and API, which represent the package name, action keywords, and API instructions of the application respectively. We regularize these words and action histories to check whether they select a sufficient and correct number of applications, UI elements, and APIs to accomplish the given task.

\begin{figure}
  \centering
    \begin{minipage}[t]{0.48\textwidth}
        \centering
        \includegraphics[width=\textwidth]{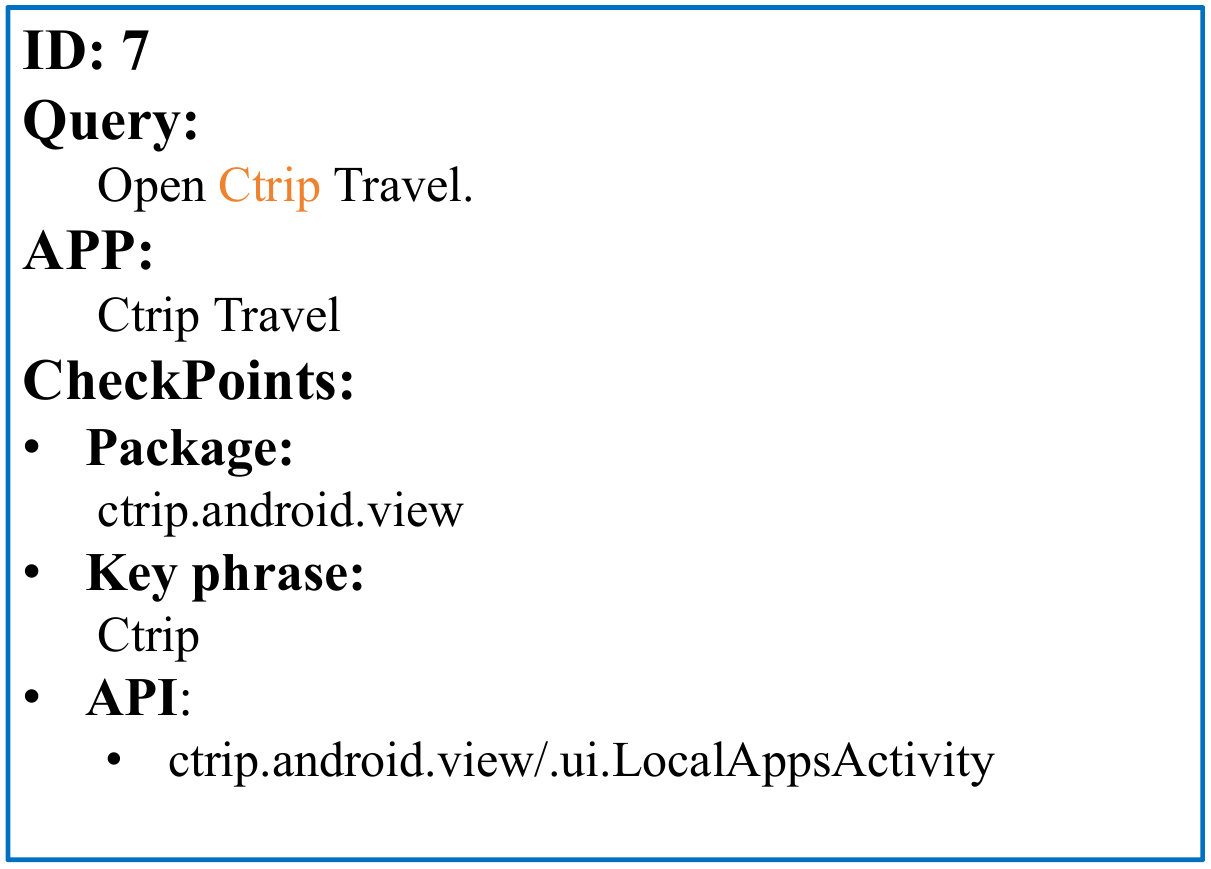}
        \caption{A test case in SAST.}
        \label{fig:case_sast}
    \end{minipage}
    \begin{minipage}[t]{0.48\textwidth}
        \centering
        \includegraphics[width=\textwidth]{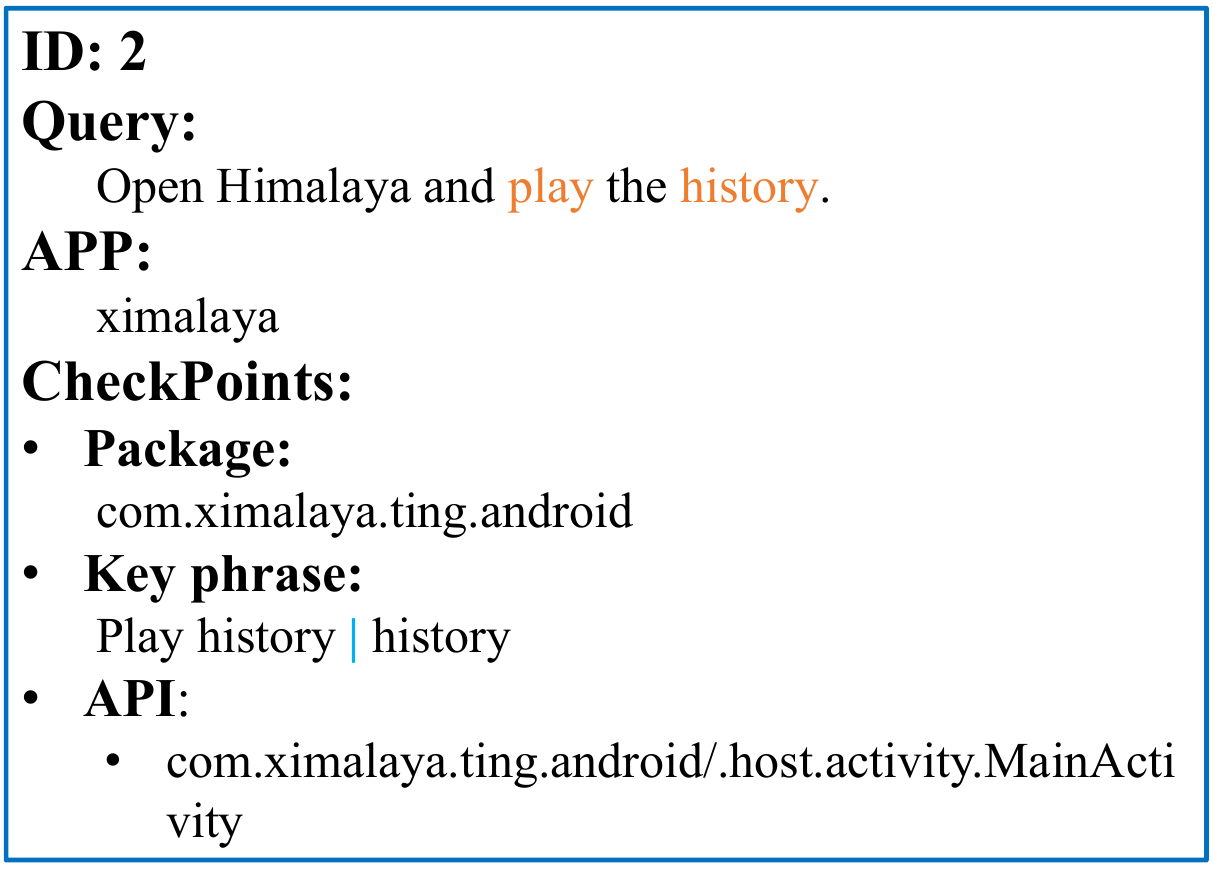}
        \caption{A test case in SAMT.}
        \label{fig:case_samt}
    \end{minipage}
\end{figure}

Next, we will give an example of CheckPoints in Figure \ref{fig:case_sast} and Figure \ref{fig:case_samt}.

\begin{figure}
  \centering
    \begin{minipage}[t]{0.48\textwidth}
        \centering
        \includegraphics[width=\textwidth]{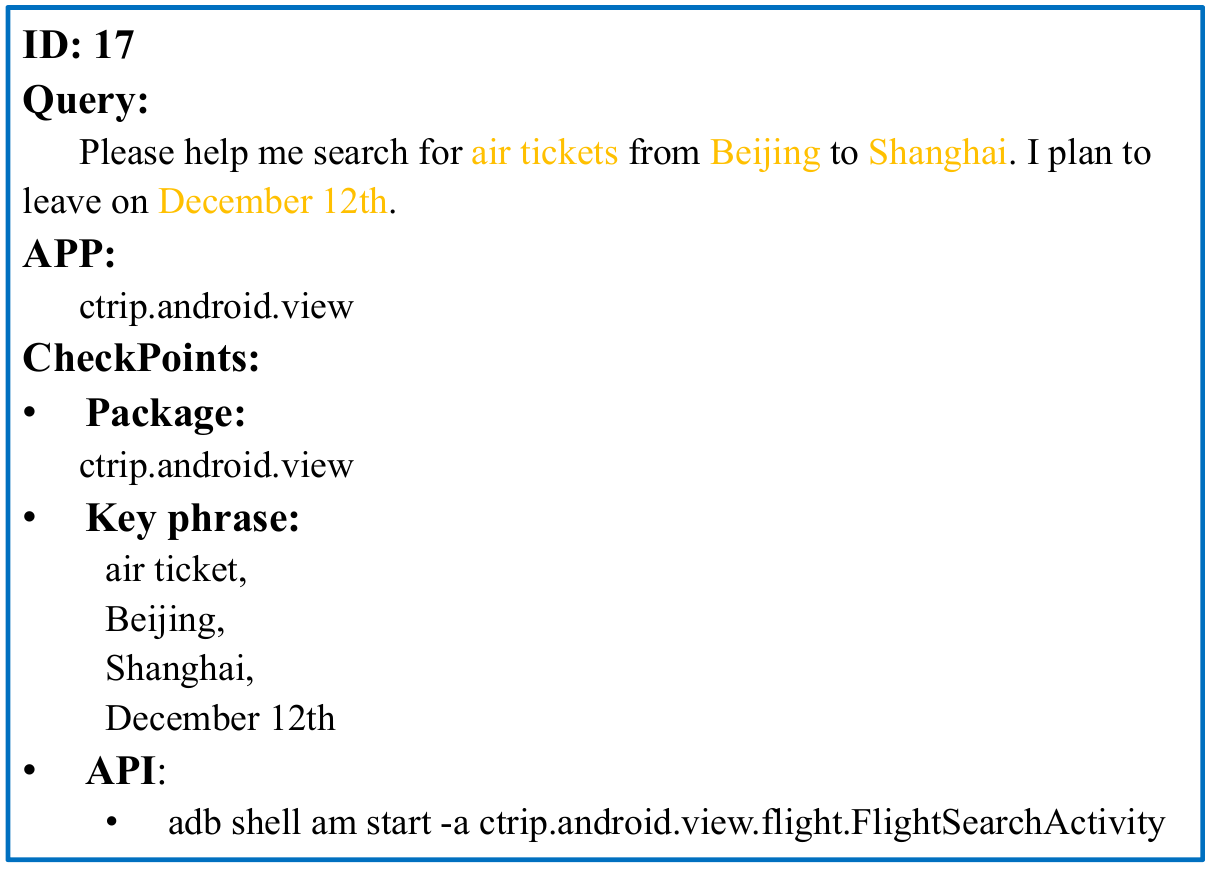}
        \caption{A test case in SAMT.}
        \label{fig:case_samt_ac}
    \end{minipage}
    \begin{minipage}[t]{0.48\textwidth}
        \centering
        \includegraphics[width=\textwidth]{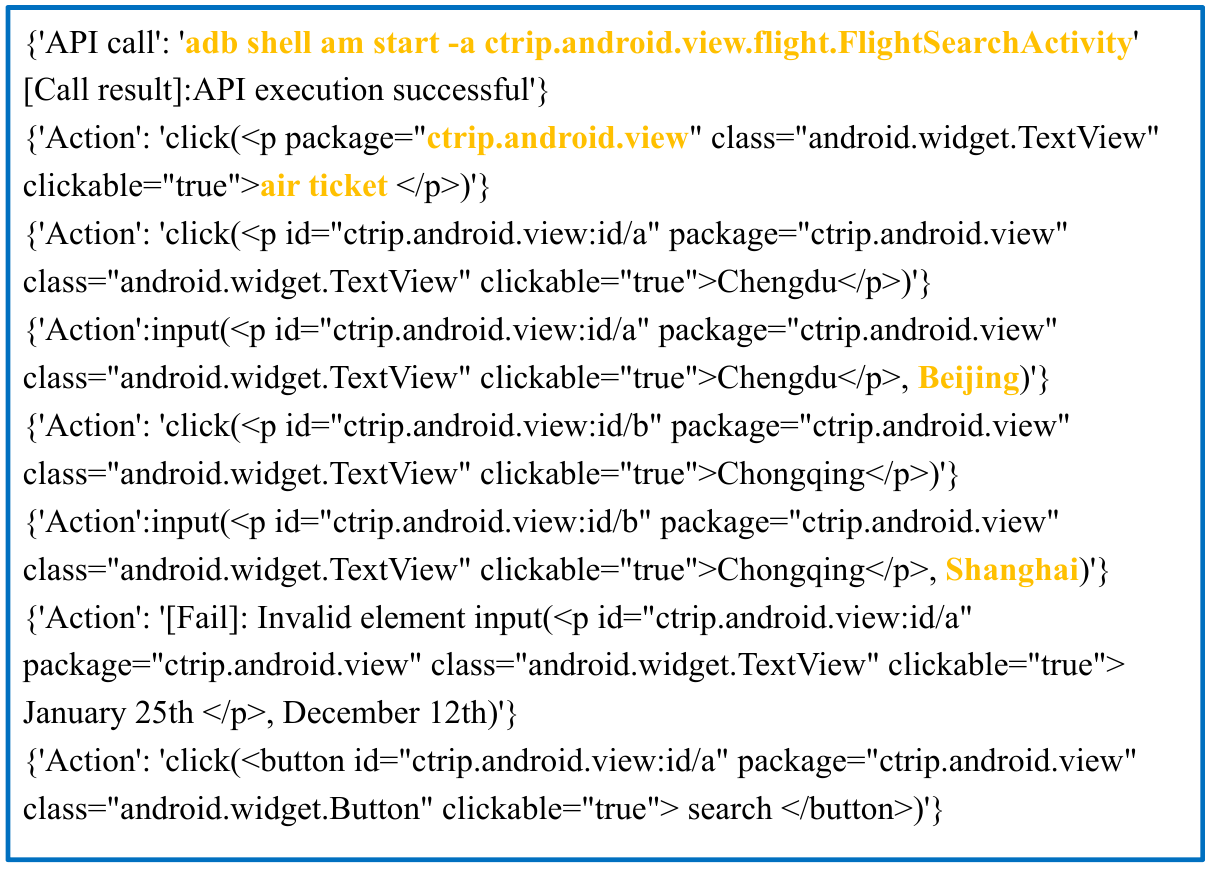}
        \caption{A action history of a test case in SAMT.}
        \label{fig:case_ac}
    \end{minipage}

\end{figure}

Figure \ref{fig:case_samt_ac} and Figure \ref{fig:case_ac} is an example of a data set and action history. Note that CheckPoints only check successfully executed instructions in the action history. From the action history, we can see that the emulator successfully opened the application by API, perform tasks in ctrip package, and selected "air ticket", "Beijing", and "Shanghai" elements, but failed to input the correct date. According to the definitions of level 1 and level 2 CheckPoints, level 1 CheckPoint score counts package CheckPoints covered, and the score of the example is 1/1, level 2 CheckPoint score counts all CheckPoints covered, and the score of the example is 5/6.

\subsection{Supplementary experiments}
As can be seen from the table \ref{tab:app_category}, categories with smaller average execution steps generally have higher success rates and CheckPoints scores. Among them, the travel transportation task has the largest average number of execution steps and the lowest PassRate. We can think that more complex tasks require longer execution steps, and the PassRate and CheckPoint score of complex tasks are lower. Travel transportation task contains more uncertainties and it is difficult to determine whether it is completed, so the PassRate is the lowest.
\begin{table*}[htbp]
\centering
\renewcommand{\arraystretch}{1.2}
\begin{adjustbox}{max width=0.85 \textwidth}
\begin{tabular}{lccccc}
\hline
APP Category & Case Quantity & Average \#Steps & PassRate(\%) & CheckPoint\textsubscript{l1} & CheckPoint\textsubscript{l2} \\
\hline
Travel Transportation & 18 & 8.17 & 39 & 83 & 68 \\
Audiovisual Vision & 34 & 4.03 & 82 & 68 & 72 \\
Social Communication & 30 & 6.40 & 77 & 57 & 63 \\
Fashion Shopping & 35&7.97 & 54 & 63 & 61 \\
Information News & 24 & 6.46 & 67 & 83 & 68 \\
Practical Tool & 61& 2.08 & 89 & 87 & 89 \\
Home Life & 46 & 1.67 & 89 & 72 & 91 \\
Book Reading &23 & 4.17 & 78 & 74 & 84 \\
Universal Buttons & 61 & 1.20 & 98 & 98 & 99 \\
\hline
\end{tabular}
\end{adjustbox}
\caption{Results on SAST classified by APP categories}
\label{tab:app_category}
\end{table*}

\section{Details for Baseline Model}
\subsection{Examples for HTML}
\begin{figure*}[th]
  \centering
        \includegraphics[width=\textwidth, page=1]{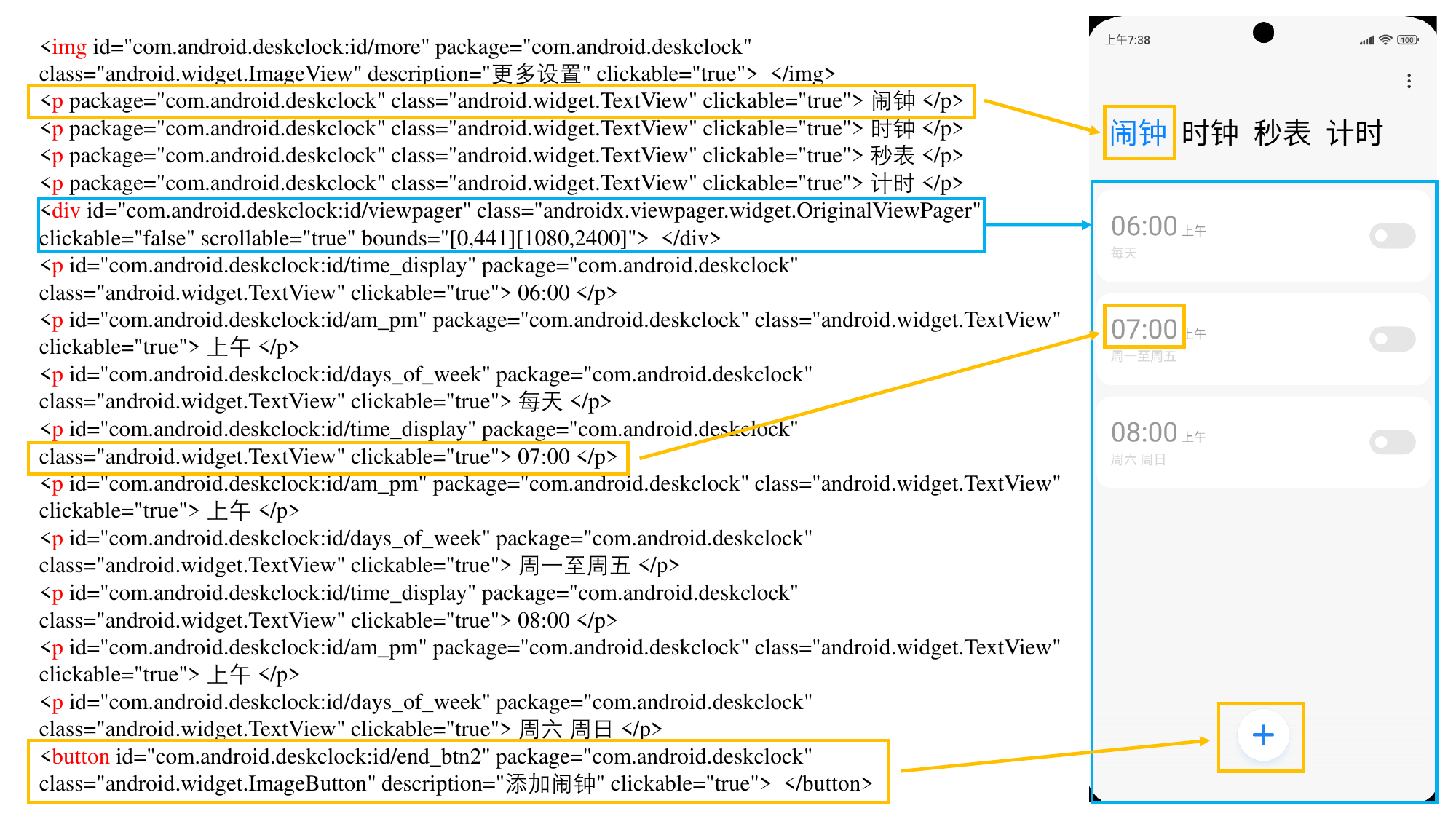}
        \caption{An example for HTML. The orange box illustrates clickable elements, and the blue frame illustrates the scrollable range.}
        \label{fig:html}
\end{figure*}
Figure \ref{fig:html} shows the correspondence between the components in the UI page and the corresponding HTML code. 
It is easy to find that most components have text descriptions, but the switch of the alarm clock does not have a corresponding text description, and LLM will hardly think of it. 
To click this button, therefore, component function exploration is what we need to do next.

\subsection{Prompts for application Selection and Planning}
You are a large language model agent stored on a mobile phone, below I will provide you with a task, the environment of the current mobile phone interface(Apps information).

Please help me choose the correct APP to perform the task based on the Apps information.
If the APP you want is not available on the current page, you can go to play store and download a suitable APP.

On this basis, you should make a simple plan for completing the task.

Let's Begin!




\subsection{Prompts for API Selection}
You are the greatest large language model agent stored on a mobile phone. You will be provided with a API list that can be called by mobile phone, the task you need to complete, the thought about what have done and what need to do now.

You are just the first step to interact with the phone, and your follow-up is UI interaction components. If you find that there is no suitable API and the next step is UI interaction, please answer directly sorry. You should not use the API to complete the work that has been completed by the UI interactive components in the previous steps.

Your decision should consider the following factors:

        \indent1. You need to first judge based on the UI information and actions complete whether the planned action has been completed.
        
        \indent2. You must only choose one API that should be executed most at present to finish the first action in next actions.
        
        \indent3. If there is no suitable API, you can just say sorry without providing any additional suggestions.
        
Strings within "$<>$" needs to be replaced with specific parameters, you must return a fully executable adb command. Perhaps you can hand over this task to the UI interaction module.

[API list]: 

[Examples]:

"adb shell input tap <x> <y>" is strictly prohibited as an answer. Your [Answer] can only follow the two templates: "Yes, the most suitable API function call is [adb command]" or "Sorry, [explain]".

Let's Begin!






\subsection{Prompts for UI Selection}
You are a large language model agent stored on a mobile phone, You need to give the current one-step action that needs to be taken to complete the task.
Below I will provide you with a task, a plan, the environment of the current mobile phone interface(UI information), action history, though about the current status of task completion. 

You need to select the most suitable one element and give the corresponding one action based on the UI information and thought.
You need to first judge based on the UI information and action history whether the planned action has been completed.
Your selection should also consider action history, and have the courage to try new buttons instead of the same buttons from history.

Action can only be the following three functions: 

    \indent 1. click(element) 
    
    \indent Click a element, only when clickable="true", the element can be clicked.
    
    \indent 2. input(element, text)
    
    \indent When you decide to enter, you first need to select the unit by clicking.
    
    \indent 3. scroll $[x_{start},y_{start}][x_{end},y_{end}]$ 
    
    \indent Scroll the screen from $[x_{start},y_{start}]$ to $[x_{end},y_{end}]$. 
    The four parameters you fill in cannot be directly the same as $x_{min}$, $y_{min}$, $x_{max}$, $y_max$. $x$ cannot exceed $(x_{min}, x_{max})$, and $y$ cannot exceed $(y_{min}, y_{max)}$.

[Examples]:

Remember: 

\indent1.Click and input have higher priority than scrolling. Scrolling is only considered when all elements of the current interface are indeed irrelevant to the task.

\indent2.When you fail to try repeatedly in one interface, maybe you can try to turn back to select other options.

\indent3.When you need to switch APPs, you need to return to the desktop first.

\indent4.When input fails multiple times, you should first select it with click.

Let's Begin!






\subsection{Prompts for Thought Generation}
You are a large language model agent stored on a mobile phone, below I will provide you with a task, a plan,
the environment of the current mobile phone interface before action (Previous UI information), current action, the environment of the current mobile phone interface(Now UI information), action history. Action history records completed operations, including click, input, scroll and API list.

You need to summarize these four aspects: changes in the UI page, actions that have been completed, task progress, one next action. 

[one next action] need to choose one among click, input, scroll and one API as the next action, and give one and only one operation object. [One next action] strictly refer to [current action] and [action history] result to do the next action.

[action history] are all previous historical actions, and [current action] is the current action that causes the UI page to change.

[Examples]:

Let's Begin!







\subsection{Prompts for Task Completion}
You are a large language model agent stored on a mobile phone, below I will provide you with a task, the environment of the current mobile phone interface(UI information), historical action information.
You need to judge whether the current task has been completed based on the current environment and historical action information.

\subsection{Algorithm Examples}
\label{Algorithm Examples}
This is a running process of the algorithm on a test case of SAMT \\
$[$ data $]$: \\
$\ \{$\\
\indent "id": 2,  \\
\indent "query": $[$ \\
\indent \indent "Play recent records in history with Himalaya." \\
\indent $]$, \\
\indent "check$\_$point": $\{$ \\
\indent\indent "activity":$[$ \\
\indent\indent\indent "com.ximalaya.ting.android.host.activity.MainActivity", \\
\indent\indent\indent $\&$ "com.ximalaya.ting.android.host.activity.MainActivity" \\
\indent\indent $]$, \\
\indent\indent "key phrase": $[$ \\
\indent\indent\indent "Playing history" | "history " \\
\indent\indent $]$, \\
\indent"package": "com.ximalaya.ting.android" \\
\indent $\}$, \\
\indent "domain": "smartApp/Ximalaya" \\
$\}$

According to algorithm \ref{alg:example}, LLM generates a plan based on the query in data as a task and a given list of available applications as follows: \\
$[$Task$]$:  Play recent records in history with Himalaya. \\
$[$app list$]$: \\
$[$ \\
\indent $\{$ \\
\indent\indent "name": "ctrip", \\
\indent\indent "function$\_$description": "As an authoritative online travel service company in the industry, Ctrip's travel hotel booking platform covers approximately 1.4 million hotels around the world; air ticket products cover more than 300 international airlines; these airlines operate flights to major cities around the world, providing more than 2 million Air routes connect more than 5,000 cities in about 200 countries and regions..." \\
\indent $\}$, \\
... \\
$]$ \\
$[$Plan$]$: I should open the Himalaya app and find the "Playback History" option in the app to view recent playback history. \\

Then the task plan is executed iteratively. In each iteration, the model accepts task, app list, action history, plan, and current HTML as input, and outputs execution API calls or UI operation instructions. 

A short example of AH(action history), thought, Html will be shown below: \\
$[$Action History$]$:  \\
$\{$'API call': 'adb shell am start -n com.ximalaya.ting.android/.host.activity.MainActivity. [Call result]:API execution successful'$\}$ \\
$\{$'Action': '$[$Fail$]$: Invalid element click(<div id="com.ximalaya.ting.android:id$/$xm$\_$ad$\_$host$\_$count$\_$down \\
$\_$click$\_$lay$\_$no$\_$click$\_$view" package="com.ximalaya.ting.android" class="android.view.View" description="Skip ads" clickable="true">  </div>)'$\}$ \\
$\{$'Action': 'click(<button id="com.ximalaya.ting.android:id/tab$\_$myspace$\_$and$\_$listen" package="com.ximalaya.ting.android" class="android.widget.RadioButton" clickable="true"> my </button>)'$\}$ \\
$[$thought$]$:  \\
Changes: The current page is the main page of the Ximalaya app. There is no change between the previous and current pages. \\
Task progress: From the current action and action history, I am currently on the main page of the Ximalaya app, ready to check the play history. \\
Actions completed: From the current action and action history, I have opened the Ximalaya app. \\
One next action: Click on the "play history" option to view the recent play history. \\
$[$HTML$]$: \\
<p package="com.ximalaya.ting.android" class="android.widget.TextView" clickable="true"> membership </p> \\
<p id="com.ximalaya.ting.android:id/main$\_$tv$\_$user$\_$level" package="com.ximalaya.ting.android" class="android.widget.TextView" clickable="true"> user level </p> \\
<p id="com.ximalaya.ting.android:id/main$\_$tv$\_$listen$\_$duration$\_$title" package="com.ximalaya.ting.android" class="android.widget.TextView" clickable="true"> Listen (minutes) </p> \\
<p package="com.ximalaya.ting.android" class="android.widget.TextView" clickable="true"> fans </p> \\
<p package="com.ximalaya.ting.android" class="android.widget.TextView" clickable="true"> Follow </p>  \\
<p package="com.ximalaya.ting.android" class="android.widget.TextView" clickable="true"> live </p>  \\
<button package="com.ximalaya.ting.android" class="android.widget.Button" clickable="true"> message </button>  \\
<button package="com.ximalaya.ting.android" class="android.widget.Button" clickable="true"> history </button>  \\
<button package="com.ximalaya.ting.android" class="android.widget.Button" clickable="true"> favorite </button>  \\
<button package="com.ximalaya.ting.android" class="android.widget.Button" clickable="true"> local </button>  \\
<button package="com.ximalaya.ting.android" class="android.widget.Button" clickable="true"> purchased </button>  \\
<img package="com.ximalaya.ting.android" class="android.widget.ImageView" description="play" clickable="true">  </img>  \\
A example of api action or ui action will be shown below: \\
$[$Action$]$: $[$adb shell am start -n com.ximalaya.ting.android/.host.activity.MainActivity$]$ \\
$[$Action$]$: click(<button package="com.ximalaya.ting.android" class="android.widget.Button" clickable="true"> history </button>)

After successful execution, the current action will be added to the action history, the updated HTML of the emulator will be read, and handed over to LLM to generate a new thought and determine whether the task is over.\\
$[$thought$]$:  \\
Changes: The current page is the "My" page in the Ximalaya app.  \\
Actions Complete: I have opened the Ximalaya app and clicked the "my" button, then clicked the "play history" button. \\
Task progress: The current mission progress is to view the play history. \\
One next action: Click on the "play" item to continue playing. \\
$[$Finished$]$: No, task is not finished.

\end{document}